\documentclass[journal,anonymous=true]{IEEEtran}
\IEEEoverridecommandlockouts

\usepackage{cite}
\usepackage{cuted}

\usepackage{mathrsfs}
\usepackage{lipsum}
\usepackage{graphicx}
\usepackage{epsfig}
\usepackage{float}
\usepackage{amsfonts}
\usepackage{dsfont}
\usepackage{subcaption}
\usepackage{setspace}
\usepackage[ruled,linesnumbered,noend]{algorithm2e}
\usepackage{algpseudocode}
\usepackage{threeparttable}

\usepackage{array}

\usepackage[utf8]{inputenc} 
\usepackage[T1]{fontenc}    
\usepackage{graphicx} \graphicspath{ {figures/} }
\usepackage{times}
\usepackage{graphicx}
\usepackage{amsmath,amssymb}
\usepackage{acronym}
\usepackage{enumitem}
\usepackage[breaklinks,colorlinks,linkcolor=black]{hyperref}
\usepackage{balance}
\usepackage{xspace}
\usepackage{setspace}
\usepackage[skip=3pt,font=small]{subcaption}
\usepackage[skip=3pt,font=small]{caption}
\usepackage[dvipsnames,svgnames,x11names]{xcolor}
\usepackage[capitalise,nameinlink]{cleveref}
\usepackage{booktabs,tabularx,colortbl,multirow,multicol,array,makecell}
\usepackage{dblfloatfix}
\usepackage[misc]{ifsym}
\usepackage{pifont}
\usepackage{cite}

\makeatletter
\DeclareRobustCommand\onedot{\futurelet\@let@token\@onedot}
\def\@onedot{\ifx\@let@token.\else.\null\fi\xspace}

\makeatother

\makeatletter
\def\BState{\State\hskip-\ALG@thistlm}
\makeatother

\makeatletter
\renewcommand{\paragraph}{%
  \@startsection{paragraph}{4}%
  {\z@}{0ex \@plus 0ex \@minus 0ex}{-1em}%
  {\hskip\parindent\normalfont\normalsize\bfseries}%
}
\makeatother

\crefname{algorithm}{Alg.}{Algs.}
\Crefname{algocf}{Algorithm}{Algorithms}
\crefname{section}{Sec.}{Secs.}
\Crefname{section}{Section}{Sections}
\crefname{table}{Tab.}{Tabs.}
\Crefname{table}{Table}{Tables}
\crefname{figure}{Fig.}{Fig.}

\definecolor{gblue}{HTML}{4285F4}
\definecolor{gred}{HTML}{DB4437}
\definecolor{ggreen}{HTML}{0F9D58}

\definecolor{mygray}{gray}{.92}

\begin{document}


\title{CoINS: Counterfactual Interactive Navigation via Skill-Aware VLM}



\author{Kangjie Zhou, Zhejia Wen, Zhiyong Zhuo, Zike Yan, Pengying Wu, Ieng Hou U, Shuaiyang Li, Han Gao, Kang Ding, Wenhan Cao, Wei Pan and Chang Liu
\thanks{Kangjie Zhou, Zhejia Wen, Zhiyong Zhuo, Pengying Wu, Ieng Hou U, Shuaiyang Li, Han Gao, Kang Ding, and Chang Liu are with School of Advanced Manufacturing and Robotics, Peking University.}
\thanks{Zike Yan is with Department of Mechanical and Automation Engineering, The Chinese University of Hong Kong.}
\thanks{Wenhan Cao is with College of Design and Engineering, National University Of Singapore.}
\thanks{Wei Pan is with Department of Computer Science, The University of Manchester.}
}
\maketitle

\begin{strip}
\vspace{-3cm}
\centering
\includegraphics[width=\linewidth]{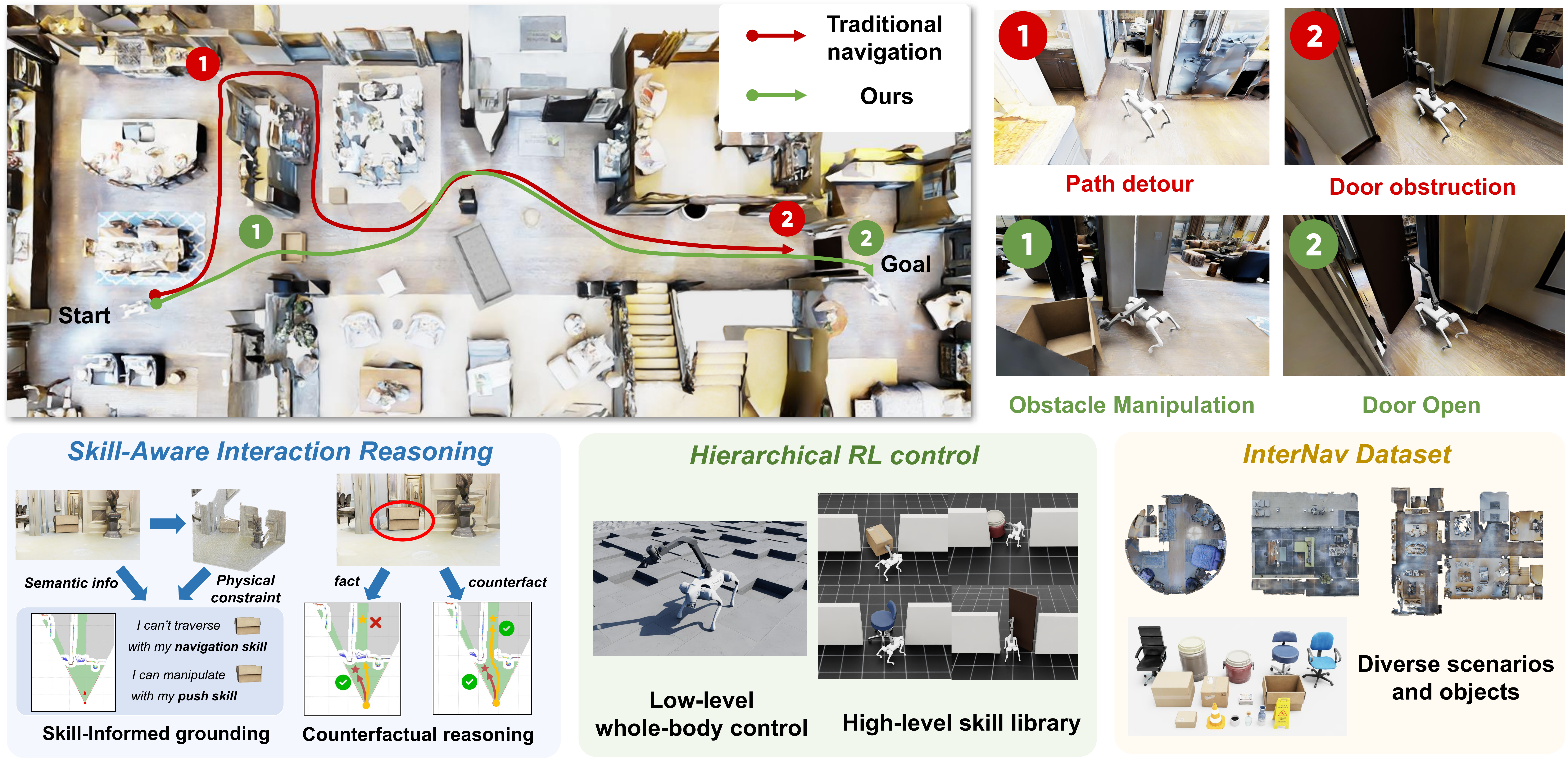} 
\captionof{figure}{Overview. We introduce CoINS, a hierarchical interactive navigation framework that integrates skill-aware VLM reasoning with RL-based skill library of diverse behaviors. Unlike traditional navigation methods that rely on passive obstacle avoidance, CoINS enables the robot to actively reconfigure the environment to clear paths in cluttered scenarios.} 
\label{Fig: cover_pic} 
\end{strip}

\begin{abstract}
Recent Vision-Language Models (VLMs) have demonstrated significant potential in robotic planning. However, they typically function as semantic reasoners, lacking an intrinsic understanding of the specific robot's physical capabilities.
This limitation is particularly critical in interactive navigation, where robots must actively modify cluttered environments to create traversable paths.
Existing VLM-based navigators are predominantly confined to passive obstacle avoidance, failing to reason about when and how to interact with objects to clear blocked paths.
To bridge this gap, we propose Counterfactual Interactive Navigation via Skill-aware VLM (CoINS), a hierarchical framework that integrates skill-aware reasoning and robust low-level execution.
Specifically, we fine-tune a VLM, named InterNav-VLM, which incorporates skill affordance and concrete constraint parameters into the input context and grounds them into a metric-scale environmental representation.
By internalizing the logic of counterfactual reasoning through fine-tuning on the proposed InterNav dataset, the model learns to implicitly evaluate the causal effects of object removal on navigation connectivity, thereby determining interaction necessity and target selection.
To execute the generated high-level plans, we develop a comprehensive skill library through reinforcement learning, specifically introducing traversability-oriented strategies to manipulate diverse objects for path clearance.
A systematic benchmark in Isaac Sim is proposed to evaluate both the reasoning and execution aspects of interactive navigation.
Extensive simulations and real-world experiments demonstrate that CoINS significantly outperforms representative baselines, achieving a 17\% higher overall success rate and over 80\% improvement in complex long-horizon scenarios compared to the best-performing baseline, while exhibiting robust generalization across diverse object categories.
Our project page is available at \url{https://coins-internav.github.io/}.
\end{abstract}


\section{Introduction}
\label{sec:introduction}

Interactive navigation extends autonomous capabilities from passive obstacle avoidance to proactively and physically modifying the environment to create traversable paths and reach the goal~\cite{stilman2005navigation,wang2023camp}. 
In contrast to conventional navigation approaches that rely on the existence of collision-free corridors~\cite{hoeller2024anymal, wellhausen2021rough}, interactive navigation addresses complex real-world scenarios where valid paths may initially be non-existent and interaction with movable obstacles becomes a prerequisite for successful navigation, such as in cluttered domestic environments and disaster response sites. 
As illustrated in the top row of \Cref{Fig: cover_pic}, failing to interact in these scenarios forces robots to take inefficient detours or renders the goal entirely unreachable when blocked by obstacles. 
In such cases, the robot is required to proactively reconfigure the environment structure through physical interactions to construct a feasible route. 
While traditional navigation among movable obstacles (NAMO) methods~\cite{ouyang2024long,bi2025interactive} attempt to address this, they are often constrained by assumptions of global map availability, which limits their applicability in partially observable navigation scenarios. Furthermore, they typically rely on simplistic geometric representations that overlook the semantic properties of objects, leading to failures in identifying objects that afford interaction within complex environments.  

Recently, vision-language models (VLMs) have been widely adopted for 
robotic navigation tasks to reason about language instructions and semantic landmarks without reliance on global maps~\cite{shah2023vint,shah2023lmnav,huang2023vlmaps,wu2024voronav,zhou2024navgpt}. 
However, existing VLM-based navigation methods are confined to passive obstacle avoidance strategies. 
When confronted with obstructions in cluttered scenarios, these models lack the causal reasoning capabilities required to reason about interaction necessity and target object selection.
Specifically, they struggle to correctly decide when to interact and which object to interact with. 
Furthermore, existing VLMs primarily function as semantic planners, lacking an intrinsic understanding of the specific robot's physical capabilities. 
For instance, a VLM might plan to cross an obstacle that exceeds the robot's traversal height or attempt to grasp a bottle beyond the robot's reach. 
We define skill-awareness as the ability to comprehend the skill semantic affordances (e.g., "climb" skill can stride a cardboard box) and evaluate its physical constraints (e.g., whether the robot can traverse over a cardboard box of a specific height with the "climb" skill).
This neglect of skill-aware understanding during the decision-making process often leads to strategies that are either suboptimal or physically infeasible for the specific robot capabilities.

To bridge this gap, we introduce a hierarchical framework named CoINS (\Cref{Fig: cover_pic}), which integrates high-level skill-aware interaction reasoning via VLM and low-level skill execution via reinforcement learning (RL) for interactive navigation. 
To endow the VLM with skill-awareness, we explicitly inject parametric skill affordances and physical constraints into the reasoning context, while grounding them within a metric-scale environmental representation. 
This approach enables the model to perform physically grounded reasoning, ensuring that generated plans are feasible for the specific robot embodiment.
To equip the VLM with interaction reasoning capabilities—specifically, determining when to interact and which object to interact with—we fine-tune the model via counterfactual reasoning, which enables the robot to implicitly evaluate the causal impact of object removal on navigation goal reachability.
Aligned with the VLM reasoning, we develop an RL-based skill library that translates the high-level skills into low-level control commands for diverse behaviors.
The main contributions of this article are summarized as follows:
\begin{itemize}
\item We propose CoINS, a hierarchical framework that enables robots to perform interactive navigation in unknown, cluttered environments without a global map, effectively integrating VLM reasoning with diverse physical interaction.

\item We develop a skill-aware VLM model, named InterNav-VLM, that internalizes counterfactual reasoning logic to determine interaction necessity and target object selection.
By integrating skill affordances and constraints into both the environmental grounding and the reasoning context, we enable the model to perform physically grounded reasoning with the specific robot capabilities.

\item To enrich the robot's physical competence, we construct an RL-based skill library. We specifically introduce a traversability-oriented manipulation paradigm, which improves navigation efficiency during blocked path clearing.

\item We introduce the InterNav dataset in Isaac Sim, featuring diverse indoor scenes and physically realistic interactive assets, which fills the gap in existing navigation datasets by providing large-scale ego-centric visual data in interactive environments for model training. 
Extensive simulations and real-world experiments demonstrate that CoINS outperforms existing baselines with higher success rate and less path length, while exhibiting robust generalization across diverse object categories.
\end{itemize}

The remainder of this paper is organized as follows: 
Section II reviews related work. 
Section III formalizes the problem and introduces the CoINS framework.
Section IV introduces InterNav-VLM, detailing its skill-aware grounding and interaction reasoning via counterfactual logic. 
Section V describes the construction of the RL-based skill library. 
Section VI introduces the InterNav dataset.
Sections VII and VIII present simulation and real-world experimental results, respectively, and Section IX concludes the paper.

\section{Related Work}

\subsection{Interactive Navigation}

Interactive navigation addresses the challenge of autonomous navigation in cluttered environments where static paths to the goal do not exist. This problem has its roots in Navigation Among Movable Obstacles (NAMO)~\cite{stilman2005navigation, stilman2008planning, yang2025efficient}, which focused on planning sequences of object manipulations to clear a path. Early NAMO approaches relied on complete global knowledge of the environment, including precise maps and object poses, treating the problem as a high-dimensional configuration space search. While theoretically sound, these methods struggle in real-world scenarios where environments are partially observable and unstructured.

Recent research has shifted towards perception-driven methods that rely on onboard sensors~\cite{zhou2025adaptive, zeng2021pushing}. Wu et al.~\cite{wu2010navigation} introduced a search-based planner for unknown environments, and Muguira et al.~\cite{muguira2023visibility} proposed a visibility-aware NAMO framework that reasons about occlusion and reachability. However, these methods typically simplify interactions to axis-aligned pushes of simple geometric primitives (e.g., cubes), limiting their applicability to complex real-world objects like furniture or doors. Recently, Schoch et al.~\cite{schoch2024sight} introduced In-Sight, a framework that combines a global planner with local interactions, utilizing estimated traversability to navigate around or interact with obstacles. Although In-Sight incorporates semantic information for traversability estimation, its interaction strategy is simplistic, primarily relying on collision-based pushing without nuanced manipulation for diverse object types.


\subsection{Vision-Language Models for Robotic Navigation}

Vision-language models have demonstrated remarkable capabilities in robotic decision-making~\cite{driess2023palm, zitkovich2023rt}. Models like PaLM-E~\cite{driess2023palm} and RT-2~\cite{zitkovich2023rt} integrate visual and textual inputs to generate low-level actions, showing promise for general-purpose robot control. In the context of navigation, VLMs have been widely adopted for instruction following and semantic goal search. Methods such as LM-Nav~\cite{shah2023lm}, ViNT~\cite{shah2023vint}, and CLIP-Nav~\cite{dorbala2022clip} leverage pre-trained vision-language representations to ground language commands into topological graphs or waypoints for navigation.

However, most existing VLM-based navigation approaches assume a passive collision-free navigation paradigm, where the robot's role is solely to traverse the environment by avoiding obstacles, rather than actively interacting with them. 
Furthermore, standard VLMs lack explicit skill-awareness. They typically rely on semantic affordances but lack the ability to reason about physical constraints, leading to infeasible decisions in tight spaces. 
Recent works like OmniVLA~\cite{hirose2025omnivla} have started to explore multi-modal action generation, but explicitly conditioning VLM reasoning on parametric skill affordances for interactive navigation remains an underexplored area. Our work bridges this gap by injecting parametric constraints into the VLM to enable physically grounded reasoning about interaction necessity.


\subsection{Learning-Based Loco-Manipulation}

Mobile manipulation combines the mobility of a base with the dexterity of a manipulator, enabling robots to interact with large-scale environments. Traditional approaches often decouple navigation and manipulation, treating the base as a transport mechanism and the arm as a separate actor. However, effective interactive navigation often requires whole-body coordination, where base motion and arm manipulation are coupled to generate sufficient force or reach~\cite{blomqvist2020go, ha2024umi, zhu2025versatile}.

Recent advances in deep reinforcement learning (DRL) have enabled the learning of such coordinated whole-body skills. Fu et al.~\cite{fu2023deep} demonstrated a unified policy for locomotion and manipulation on quadrupedal robots, while Cheng et al.~\cite{cheng2023legs} explored using legs as manipulators to push objects. These works highlight the potential of end-to-end learning for achieving tight coordination between the base and manipulator. Beyond quadrupeds, Mobile ALOHA~\cite{fu2024mobile} has shown that imitation learning can acquire complex bimanual mobile manipulation skills with remarkable dexterity.

DRL has demonstrated efficacy in diverse manipulation tasks, from object goal pushing~\cite{dadiotis2025dynamic, jeon2023learning}, grasping~\cite{zeng2018learning, kalashnikov2018scalable, liu2024visual}, and contact-rich door opening~\cite{sleiman2023versatile, zhang2024learning}. Yet, a significant limitation of current learning-based methods is their limited generalization across object categories. Most policies are trained on specific object instances or simplified geometric shapes. Recent works have attempted to address this through category-level manipulation~\cite{mu2021maniskill} and large-scale skill learning~\cite{dalal2024plan}, but robust generalization remains an open challenge.

Moreover, the objective of manipulation in interactive navigation differs fundamentally from standard rearrangement tasks. Instead of precise object placement, the goal is to clear a path for the robot to pass, rather than positioning the object at a specific coordinate. We address challenges by developing the traversability-oriented manipulation policy, which prioritizes path clearance over precise object positioning.

\section{Problem Formulation and Overview of CoINS}

\begin{figure*}[!t] 
    \centering
    \includegraphics[width=\linewidth]{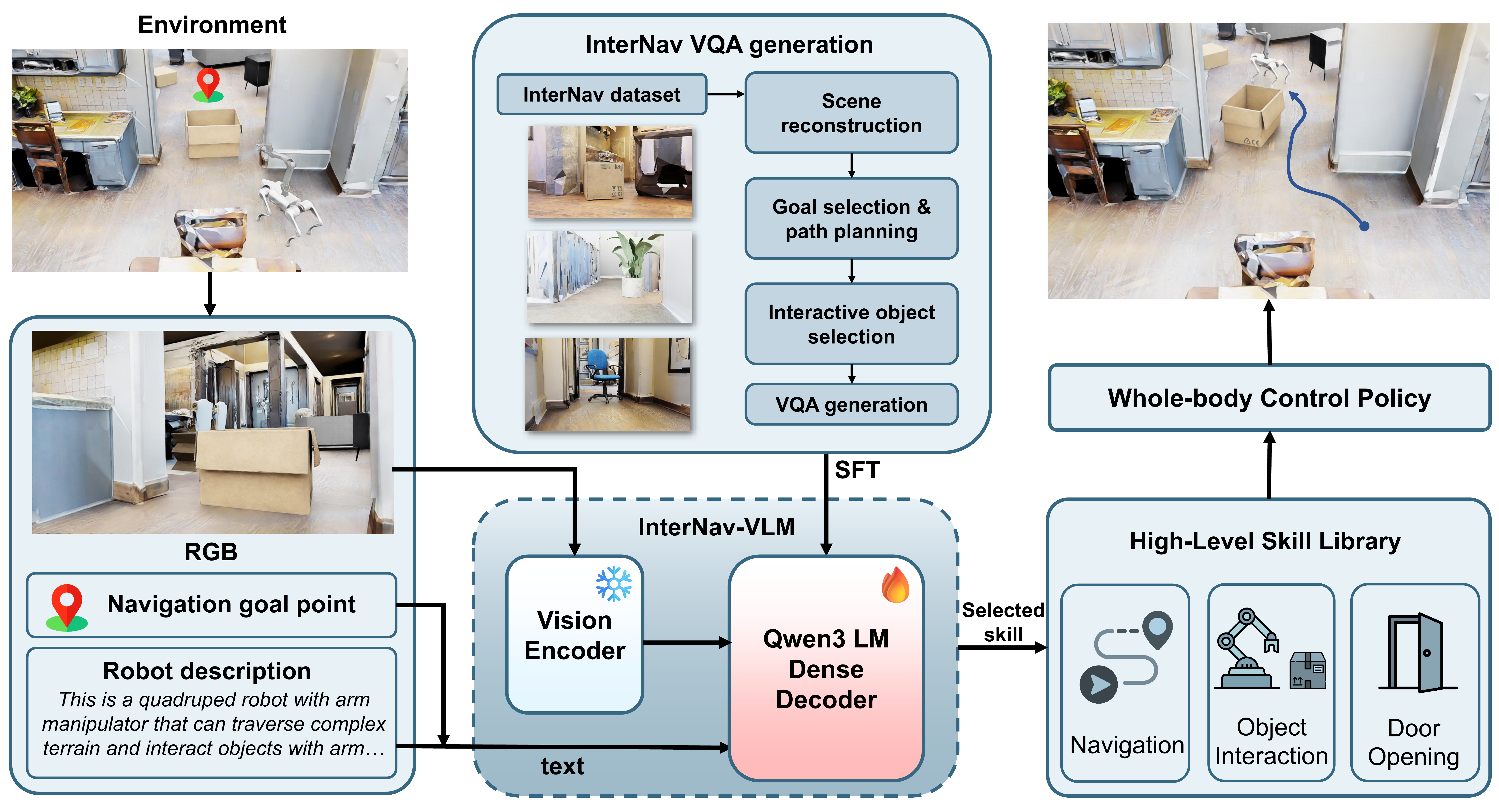} 
    \caption{InterNav-VLM in CoINS framework. The VLM reasoning module takes the robot's egocentric RGB observations and embodiment constraints as input to produce high-level interaction and navigation decisions, which are then translated by the skill execution module into precise motion controls for diverse interaction primitives.}
    \label{Fig: vlm_nav_framework} 
    \end{figure*}

We focus on interactive navigation in cluttered indoor environments, where traversable paths are frequently obstructed by various objects, as commonly encountered in residential, office, and warehouse environments.
We formulate this as a task and motion planning (TAMP) problem~\cite{garrett2021integrated}, defined by a 7-tuple $\mathcal{T} = (X_{init}, X_{goal}, \mathcal{O}, \mathcal{E}_f, \mathcal{E}_m, \mathcal{S}, \mathcal{C})$. 
Here, $X_{init}$ and $X_{goal}$ denote the robot's initial configuration and the target goal position, respectively. 
We assume that the robot has access to its current pose.
$\mathcal{O}$ denotes the robot's observation space derived solely from onboard sensors, consisting of RGB images $o^{rgb}$ and geometric data $o^{geo}$ (e.g., depth or point cloud).
The environment contains two types of entities: fixed obstacles $\mathcal{E}_f$, which are strictly static, and movable obstacles $\mathcal{E}_m$, which the robot can physically manipulate.
$\mathcal{S}$ represents the robot's symbolic skill set, defined as a collection of discrete skill names executable by the robot's embodiment.
$\mathcal{C}$ represents the robot's capabilities, encompassing both semantic affordances and parametric constraints. 

Given $X_{init}$ and $X_{goal}$, our objective is to find a hierarchical policy $\pi = \{\pi_H, \pi_L\}$ that navigates the robot to the goal in a partially observable environment.
The high-level reasoning policy $\pi_H$ outputs the optimal skill $s_t \in \mathcal{S}$ and execution parameters $q_t$:
\begin{equation}
(s_t, q_t) = \pi_H(X_{goal}, o^{rgb}_t, \mathcal{S}, \mathcal{C}),
\end{equation}
where $o^{rgb}_t$ denotes the current RGB observation.
Subsequently, the low-level execution policy $\pi_L$ generates the joint control commands $u_t$:
\begin{equation}
u_t = \pi_L(s_t, q_t, o^{rgb}_t, o^{geo}_t).
\end{equation}
Crucially, this entire process relies solely on onboard sensors and internal states, without access to the global map. 

To address this problem, we propose CoINS that consists of two key components:
(1) InterNav-VLM i.e., $\pi_H$, a VLM fine-tuned via counterfactual reasoning to serve as the high-level reasoner, which interprets egocentric RGB observations and robot capabilities to make interaction and navigation decisions;
(2) Skill library , i.e., $\pi_L$, a collection of RL-based execution policies that translate high-level decisions into motion controls for diverse navigation and interaction behaviors.
This synergistic design enables the robot to intelligently reason about complex environments while maintaining the agility to physically modify them.

\section{InterNav-VLM for Interactive Navigation Reasoning}

To endow the robot with intrinsic capability understanding and active interaction reasoning, we propose InterNav-VLM, a reasoning model that incorporates skill awareness into the decision-making loop.
By grounding reasoning in the robot's skill capabilities, the model answers two fundamental questions:
\begin{itemize}
\item \textbf{When is interaction necessary?} The model evaluates navigation feasibility by matching the terrain complexity against the robot's traversability skills to determine if environmental reconfiguration is required.
\item \textbf{Which object should be interacted with?} Among multiple candidates, the model selects the object that is both manipulable by the robot's skill set and whose removal optimally facilitates navigation progress.
\end{itemize}
In the following contents, we elaborate on the three main components of the InterNav-VLM training pipeline: skill-informed environmental grounding, interaction reasoning with counterfactual logic, and visual question answering (VQA) generation with introspective verification.

\subsection{VLM Architecture with Parametric Skill Affordances}

InterNav-VLM processes the RGB observation $o_{rgb}$ via a vision encoder and concatenates the resulting visual tokens with the textual embeddings of the navigation goal $X_{goal}$ and robot descriptions including skill set $\mathcal{S}$ and corresponding capabilities $\mathcal{C}$, as illustrated in \Cref{Fig: vlm_nav_framework}.
This multi-modal sequence is fed into an LLM backbone to autoregressively generate the final skill selection $s \in \mathcal{S}$ and the target interaction object $e \in \mathcal{E}_m$ if an interaction skill is chosen.
To mitigate the limitation of existing VLM-based planners that primarily rely on semantic affordances while neglecting critical parametric constraints, we propose incorporating parametric skill affordances directly into the system prompt to endow the foundation model with physically grounded reasoning capabilities.
Specifically, we articulate the traversability constraints including traversal height and clearance width limits for locomotion skills.
For manipulation skills, we specify the manipulable object categories and interaction range limits.
This design transforms the skill set from abstract semantic tokens into embodied, metric-aware knowledge, enabling the foundation model to perform physically grounded reasoning.
By incorporating the metric constraints as input, the model can understand not just what can be interacted with, but whether the interaction is feasible given the specific robot embodiment and environmental geometry.
\subsection{Skill-Informed Environmental Grounding}

To effectively utilize the integrated skill knowledge, the model must ground these parametric skill descriptions into the environment. We propose a skill-informed environmental grounding pipeline that reconstructs the scene in metric scale and generates skill-conditioned maps.

\textbf{Metric-Scale 3D Scene Reconstruction.}
Accurate interaction requires mapping visual observations to physical space with absolute scale.
First, we utilize VGGT~\cite{wang2025vggt} and Map-Anything~\cite{keetha2025mapanything} to estimate metric-scale depth maps and camera poses from the input RGB images.
Using the estimated camera poses, we back-project the metric depth maps into 3D point clouds.
Subsequently, we perform point cloud normalization to align the scene with a canonical coordinate frame. 
Detailed reconstruction process is provided in \Cref{appendix: scene-reconstruction}.

\textbf{Skill-Aware Spatial Grounding.}
We explicitly ground the skill affordances into the physical environment by generating a traversability map for locomotion and identifying reachable instances for manipulation. 
Based on the reconstructed scene, we generate a traversability map that explicitly incorporates the robot's skill capabilities.
Given the reconstructed point cloud $P = \{p_i = (x_i, y_i, z_i)\}$, we project it onto a 2D grid map $M$. For each grid cell $(u,v)$, we compute the maximum height relative to the local ground plane:
\begin{equation}
H(u,v) = \max \{z_i \mid (x_i, y_i) \in \text{cell}(u,v) \}.
\end{equation}
The raw occupancy map $M_{occ}$ is determined by comparing the terrain height against the robot's traversal height capability $h_{max}$:
\begin{equation}
M_{occ}(u,v) = \mathbb{I}(H(u,v) > h_{max}),
\end{equation}
where $\mathbb{I}(\cdot)$ is the indicator function.
To ensure safety, we generate the final traversability map $M_{trav}$ by inflating the obstacles with the robot's clearance radius $r_{clear} = w_{clear}/2$. This operation is formalized as:
\begin{equation}
M_{trav}(u,v) = 1 - \max_{(i,j) \in \mathcal{D}(u,v, r_{clear})} M_{occ}(i,j),
\end{equation}
where $\mathcal{D}(u,v, r_{clear})$ denotes the circular neighborhood centered at $(u,v)$ with radius $r_{clear}$. Here, $M_{trav}(u,v)=1$ indicates a traversable cell.

For manipulation skills, we map the skill-specified object categories $c_{obj}$ to physical instances.
We first obtain the bounding box $b_k$ of object $k$ via Grounding DINO~\cite{liu2024grounding} given $c_{obj}$.
We determine the object's 3D position $P_{obj}^k$ by directly retrieving the 3D point from the reconstructed point cloud $P$ that corresponds to the center pixel $(u_k, v_k)$ of the detected bounding box $b_k$:
\begin{equation}
P_{obj}^k = P_{(u_k, v_k)},
\end{equation}
where $P_{(u_k, v_k)}$ denotes the 3D coordinates of the point in the dense point cloud corresponding to the image pixel $(u_k, v_k)$.
The manipulability of object $k$ is evaluated by checking if it falls within the robot's manipulation workspace $\mathcal{W}(p)$ from at least one valid robot configuration $p$ in the traversable region:
\begin{equation}
\mathcal{F}_{manip}(k) = \mathbb{I}\left( \exists p \in \mathcal{M}_{trav} \text{ s.t. } P_{obj}^k \in \mathcal{W}(p) \right),
\end{equation}
where $\mathcal{M}_{trav}$ is the set of traversable cells, and $\mathcal{W}(p)$ represents the reachable workspace of the manipulator when the robot base is located at $p$.
This grounding process ensures that the reasoning model perceives the environment considering the robot's skill capabilities, actively filtering out physically infeasible skills.

\subsection{Interaction Learning with Counterfactual Reasoning}

A key challenge in interactive navigation is reasoning about the consequences of interactions.
We formalize this process as counterfactual reasoning: \\\textit{\ \ \ \ \ \ ``If the object were removed, would the goal become reachable?''}\\
While performing this analysis online is computationally expensive, we propose to distill this counterfactual logic into the VLM.

We first sample navigation goals and employ the A* algorithm with the skill-conditioned map $M_{trav}$ to search for feasible paths.
For each target $x_g$, we conduct reasoning to identify whether to interact and which objects to interact with.
We formalize the interaction decision as a counterfactual reasoning problem, aiming to maximize the potential navigation gain $G(o)$ achieved by removing a candidate object $o \in \mathcal{K}_{manip}$, where $\mathcal{K}_{manip} = \{o \mid \mathcal{F}_{manip}(o) = 1\}$ denotes the set of manipulable objects.
Let $l(M, x_g)$ denote the path length to the goal $x_g$ on map $M$, where $l(M, x_g)=\infty$ if the goal is unreachable.
We define the counterfactual gain as the reduction in path length:
\begin{equation}
G(o) = 1 - \frac{l(M_{trav}^{-o}, x_g)}{l(M_{trav}, x_g)},
\end{equation}
where $M_{trav}^{-o}$ denotes the counterfactual traversability map where the region occupied by object $o$ is treated as free space.
The optimal interaction target $o^*$ is selected to maximize this gain:
\begin{equation}
o^* = \operatorname*{argmax}_{o \in \mathcal{K}_{manip}} G(o).
\end{equation}

To derive the counterfactual traversability map $M_{trav}^{-o}$ with object $o$ removed, we first detect potential interactive objects using Grounding DINO~\cite{liu2024grounding} and SAM 2~\cite{ravi2024sam} to identify the mask of the object.
Then, we back-project the 2D mask into 3D space using the depth image to obtain the object's point cloud.
Subsequently, we artificially remove the object's point cloud and construct the counterfactual traversability map $M_{trav}^{-o}$ following the same procedure as the initial map generation.
If the maximum gain $G(o^*)$ exceeds a predefined threshold $\epsilon$, the interaction with object $o^*$ is deemed effective for facilitating navigation, and $o^*$ is selected as the interaction target.
Specifically, a significant gain $G(o^*) > \epsilon$ indicates that removing the object either transforms an initially infeasible path into a feasible one or substantially improves path efficiency.
Through training on this data, InterNav-VLM learns to internalize the counterfactual logic, enabling it to implicitly reason about the causal effects of interactions on navigation directly from visual inputs, without relying on computationally expensive online simulations.

\subsection{VQA Generation with Introspective Verification}

\begin{figure}[!t] 
\centering
\includegraphics[width=\linewidth]{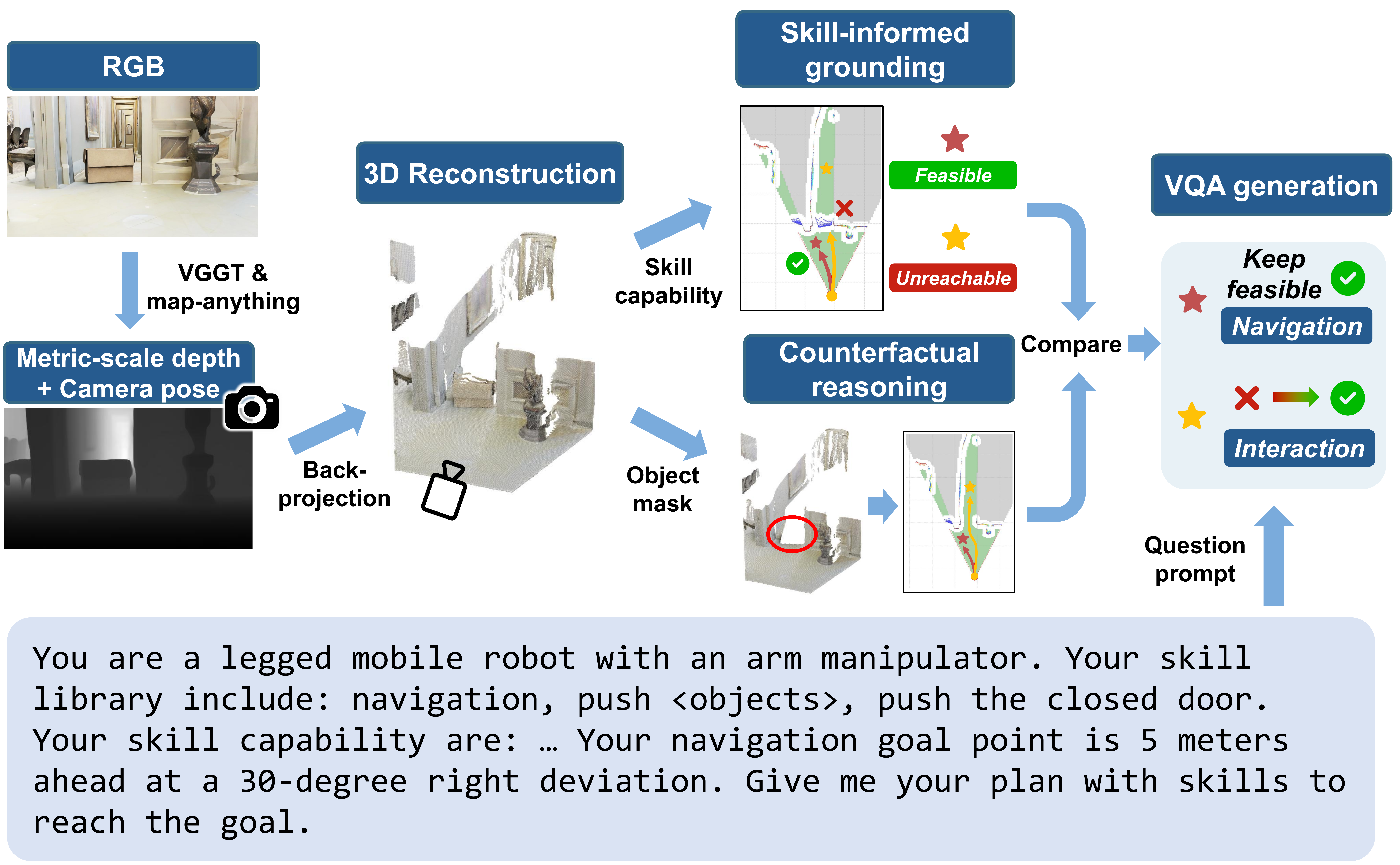} 
\caption{VQA generation process.} 
\label{Fig: VQA_generation} 
\end{figure}

To train the VLM with the aforementioned reasoning logic, we generate VQA samples for each specific target point and skill capability description by analyzing the previous path planning and interaction reasoning results.
Specifically, if the target is directly reachable, the answer corresponds to navigation. If an optimal interaction object is identified, the answer specifies the selected object along with the appropriate interaction skill.
Crucially, we integrate an introspective verification mechanism into the generation process.
Instead of directly predicting the action, we explicitly structure the chain-of-thought (CoT) reasoning that consists of two phases:
(1) Skill feasibility check: The model first verifies whether the potential action aligns with the robot's injected skill capabilities (e.g., confirming the object height is within the traversal height capability).
(2) Interaction necessity reasoning: The model then articulates the causal rationale for interaction, explicitly stating that the selected object obstructs the path and its removal is beneficial for navigation.
This introspective process acts as a soft constraint, effectively filtering out hallucinations and ensuring the generated plans are both executable and effective.
We collect approximately 20K VQA samples based on this pipeline, as illustrated in \Cref{Fig: VQA_generation}.

\subsection{Supervised Fine-Tuning}

We adopt Qwen3-VL as our backbone model to develop InterNav-VLM by fine-tuning on the generated VQA dataset to learn the interaction reasoning logic.
The model takes the RGB image $o_{rgb}$, navigation goal $x_g$, skill set $\mathcal{S}$, and embodiment capabilities $\mathcal{C}$ as inputs, and autoregressively generates the natural language answer.
The output specifically includes: (1) the selected skill from the library, and (2) the target object identifier if an interaction skill is chosen.
The training objective is the standard next-token prediction loss:
\begin{equation}
\mathcal{L} = -\sum_{i=1}^{L} \log P(y_i | y_{<i}, o_{rgb}, X_{goal}, \mathcal{S}, \mathcal{C}),
\end{equation}
where $y_i$ represents the $i$-th token in the ground-truth answer, and $y_{<i}$ denotes the sequence of preceding tokens $y_{1}, \dots, y_{i-1}$.
Through this fine-tuning process, InterNav-VLM acquires the critical reasoning capabilities required for interactive navigation, enabling it to make informed decisions on optimal skill selection based on the robot's embodiment capabilities and the environmental observations.

\section{RL-Based Skill Library for Interactive Navigation}

\begin{figure}[!t] 
    \centering
    \includegraphics[width=\linewidth]{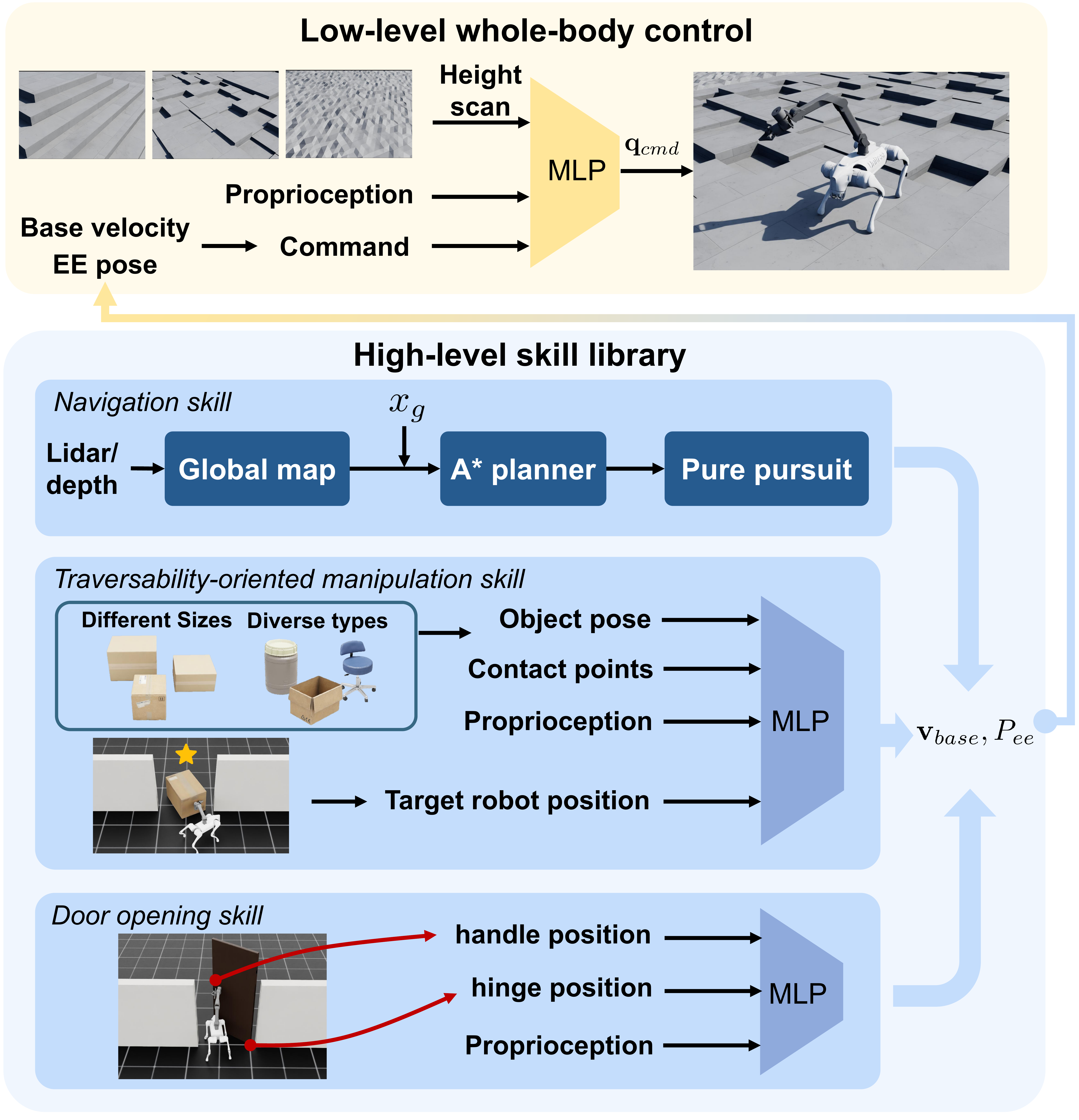} 
    \caption{RL-based hierarchical skill library.}
    \label{Fig: skill_library} 
    \end{figure}

To execute the specific skills proposed by InterNav-VLM, we employ a quadruped robot equipped with a manipulator as the embodiment, chosen for its versatile mobility and manipulation capabilities.
We develop an RL-based skill library built upon a hierarchical control framework, as shown in \Cref{Fig: skill_library}.
This framework consists of a low-level whole-body controller that ensures stable locomotion and manipulation execution, and a set of high-level skills that generate task-specific commands for the low-level controller.

\subsection{Low-Level Whole-Body Control}

We first train a robust low-level policy to handle the complex dynamics of the quadruped robot equipped with an arm-based manipulator.
This controller takes the high-level commands (the desired base velocity $\mathbf{v}_{base} = (v_x, v_y, \omega_z)$ and the target end-effector pose $P_{ee} = (p_{ee}, q_{ee})$ relative to the robot base), robot proprioception, and height scan information as inputs. It outputs the target joint positions for all 18 degrees of freedom (12 for the quadruped legs and 6 for the manipulator arm): $\mathbf{q}_{cmd} \in \mathbb{R}^{18}$.
This low-level policy is trained using Proximal Policy Optimization (PPO)~\cite{schulman2017proximal} to track the input commands while maintaining robot stability in diverse terrain.
Detailed reward function is provided in Appendix~\ref{appendix: reward-design}.

\subsection{High-Level Skill Library}

Building upon the low-level controller, we develop specialized high-level policies for navigation and diverse object interactions, where the interaction skills are learned via RL.
These policies observe the environment state and output the command sequence $(\mathbf{v}_{base}, P_{ee})$ to the low-level controller to complete specific tasks.

\textbf{Navigation Skill.}
The navigation policy enables the robot to move to a target location $x_g$ while avoiding static obstacles.
We utilize a standard navigation stack where a local map is built online using depth images and robot pose.
Given a target $x_g$, the A* algorithm generates a collision-free path in the local map.
A pure pursuit controller then tracks this path, outputting the desired base velocity $\mathbf{v}_{base}$ to the low-level controller, while the manipulator is commanded to maintain a fixed configuration to ensure stability.
This skill serves as the default behavior when no interaction is required.

\textbf{Traversability-Oriented Manipulation Skill.}
Previous approaches utilizing quadruped robots with manipulators for object interaction typically focused on pushing objects to precise target locations. However, applying these methods to interactive navigation faces two critical limitations:
(1) \textit{Goal Inconsistency}: The primary objective in interactive navigation is to clear a path for the robot to pass, rather than positioning the object at a specific coordinate.
(2) \textit{Limited Generalization}: Existing works are often restricted to simple object types like boxes and barrels, failing to handle the diverse obstacles encountered in realistic environments.

To address these issues, we propose a traversability-oriented manipulation policy.
Unlike traditional object manipulation tasks where the goal is defined by the object's final pose, our policy is robot-centric: the objective is for the robot itself to reach a target location $P_t$.
The manipulator is utilized as a tool to push obstructing objects aside, only when necessary to clear the path.
The policy learns to apply sufficient force to move the obstacle just enough to create a traversable gap, prioritizing the robot's successful arrival at the goal over the precise final configuration of the obstacle.
Formally, the objective of the traversability-oriented manipulation policy is to maximize the reward:
\begin{equation}
r = r_{nav} + r_{safe} + r_{eff},
\end{equation}
where $r_{nav}$ rewards progress towards the target location $P_t$, $r_{safe}$ penalizes collisions between the robot body and the obstacle, and $r_{eff}$ encourages energy-efficient manipulation.
To achieve robust generalization across different object types and sizes, we employ extensive domain randomization and curriculum learning during training.

The policy input includes the robot's proprioception, the relative pose of the target object, and candidate contact points sampled from the object's surface to adapt to various object geometries. The action space is the command $(\mathbf{v}_{base}, P_{ee})$ for the low-level controller.
The policy aims to reach the target location while avoiding collisions between the robot body and the obstacle, ensuring the robot's safety and stability during interaction.
During training, we randomize object properties to cover a broad spectrum of potential obstacles. This diversity primarily encompasses size variation and category diversity, including cardboard boxes, small buckets, and office chairs.
This diversity ensures the policy can adapt to the specific physical properties of the object identified by the VLM.

\textbf{Door Opening Skill.}
This skill is designed for interacting with simplified door models that can be pushed open directly, abstracting away the complexity of handle manipulation.
It takes the relative position of the door and door handle, along with the robot's proprioception, as inputs.
The training objective is to open the door to a specific angle while ensuring the robot body avoids collision with the door structure.
To achieve robust performance, we train on two door types (left-hinged and right-hinged) and randomize the door's width and height during training.

All high-level interaction skills are trained using PPO~\cite{schulman2017proximal}.
Detailed reward formulation of high-level interaction skills is provided in Appendix~\ref{appendix: reward-design}.

\subsection{Skill Execution and Parameter Specification}

During the training phase, we leverage privileged information from the simulator, providing the policy with accurate ground-truth observations of the object's state to ensure stable learning.
In the execution phase, to integrate the learned policy with the high-level VLM decision-making, we employ a vision-based pipeline to estimate the necessary state information.
When the InterNav-VLM selects an interaction skill and a target object, we convert these symbolic outputs into executable parameters for the hierarchical control system.
If the manipulation skill is selected, we first obtain the target object's pixel location from the VLM output. 
We then employ SAM 2~\cite{ravi2024sam} prompted by the pixel location to generate a precise segmentation mask. The masked points are subsequently back-projected into 3D space using the metric depth map to estimate the object's relative position.
For contact point generation, we sample points from the object's surface point cloud, derived from the back-projected mask, to serve as candidate contact locations for the manipulation policy.
The robot's local target $P_{t}$ is determined by extending the robot-to-object vector to a collision-free point, thereby guiding the robot to push through the blockage.
If the door opening skill is selected, we employ Molmo~\cite{molmo2024} to identify the 2D keypoints of the door handle and hinges in the image, which are then back-projected into 3D space using depth information to guide the manipulation policy.

Skill execution terminates when specific success criteria are met (e.g., the robot successfully passes the obstructed region, or the door is opened beyond a threshold angle). If a timeout constraint is violated, the system attempts to re-execute the skill, with the task considered failed after a predefined number of consecutive failures.

\section{Interactive Navigation Dataset}

\begin{figure}[!t]
    \centering
    \includegraphics[width=\linewidth]{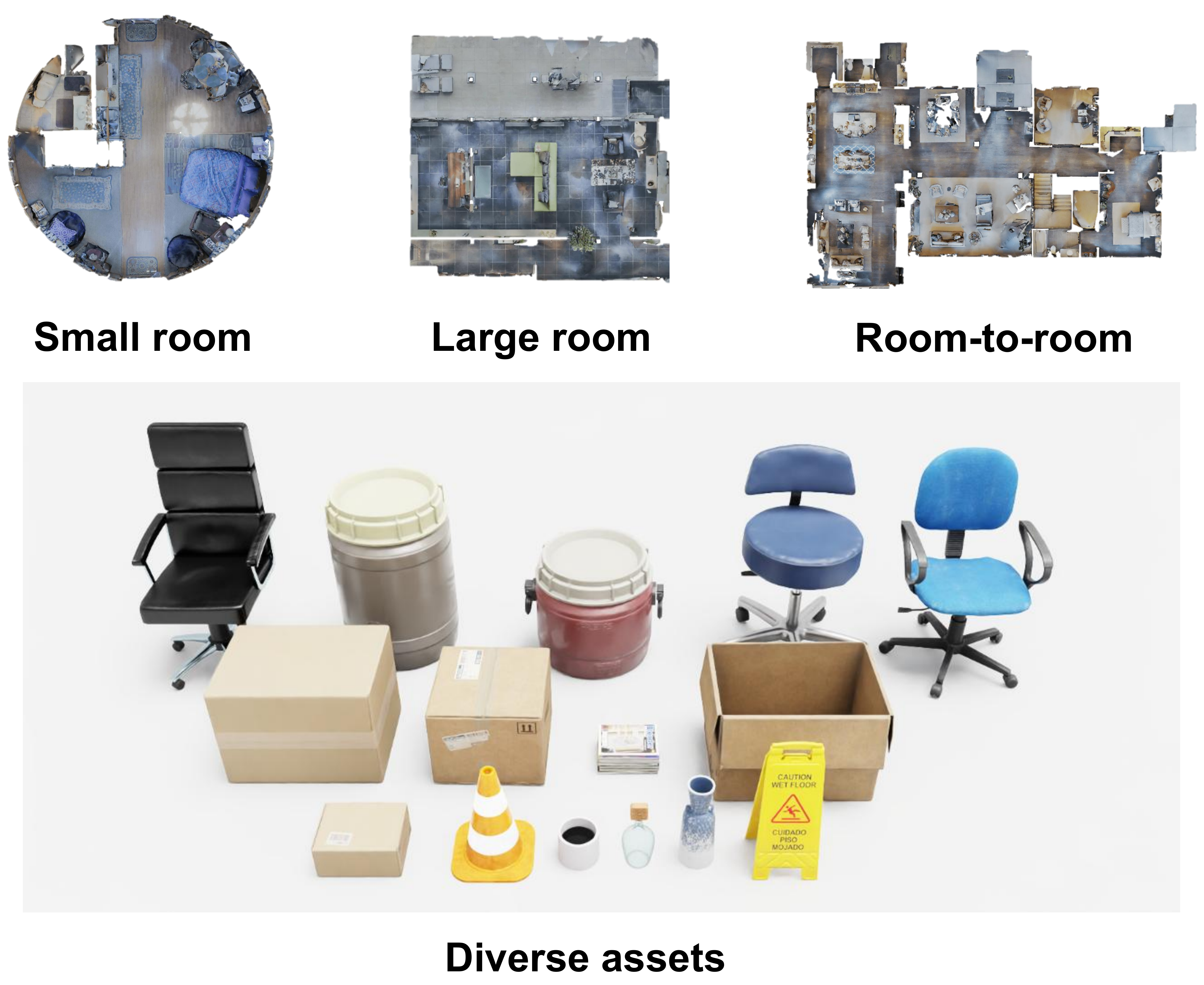}
    \caption{Diverse scenes and objects in the dataset.}
    \label{Fig: benchmark}
    \end{figure}

We introduce the InterNav dataset, a platform built on Isaac Sim to facilitate research on interactive navigation.
Unlike conventional navigation datasets that assume static environments, our dataset focuses on scenarios where interaction is essential for traversability.
The dataset serves a dual purpose: providing a diverse collection of egocentric visual data for training the VLM's reasoning capabilities on navigation and interaction, and offering a standardized benchmark with physically realistic objects to evaluate algorithm performance.
As illustrated in \Cref{Fig: benchmark}, the dataset features diverse scene layouts populated with rich assets to simulate complex real-world environments.

\subsection{Scenario Construction and Data Collection}

To acquire training data specifically tailored for interactive navigation in cluttered environments, we establish a procedural generation pipeline to collect images in diverse and cluttered simulated scenes.
Addressing the limitation that existing static navigation datasets do not support interaction, we leverage Isaac Sim to construct interactive environments.
We import scene layouts from Matterport3D~\cite{chang2017matterport3d} and procedurally populate them with a diverse set of interactable assets to create cluttered indoor scenarios.
Following scene generation, we automate the image collection process to gather raw visual data for VLM fine-tuning.
Our data collection process is characterized by three key dimensions of diversity:
\begin{itemize}
\item \textbf{Scene diversity}: Focusing on diverse indoor environments such as residential rooms, offices, and warehouses, initialized with varying clutter levels and obstacle configurations.
\item \textbf{Object diversity}: Including boxes, barrels, chairs, and doors with randomized sizes, poses, and texture materials.
\item \textbf{Viewpoint diversity}: Capturing observations with randomized camera poses, including yaw/pitch angles and mounting heights. This extensive variation simulates diverse heading directions, camera shakes, and visual perspectives from different robot embodiments.
\end{itemize}
Through this pipeline, we collect extensive data from diverse scenes and multiple perspectives, thereby enabling robust generalization across different robot platforms and viewing conditions. The collected dataset is partitioned into a training set (90\%) for model fine-tuning and a test set (10\%) for the VLM reasoning evaluation.

\subsection{Benchmark}

\textbf{Scenarios.} 
We selected diverse scenes from the Matterport3D dataset to serve as static environments, into which we imported interactive assets to construct the navigation challenges. 
The benchmark consists of 15 diverse scenes across three progressively complex categories: small room, large room, and room-to-room, each containing 5 variations. As complexity grows, the start-to-goal distance increases and the number of movable obstacles rises.
Notably, the room-to-room scenario involves situations that require opening doors.
We sample 10 start-goal pairs for each scene, resulting in a total of 150 episodes. 
Each episode concludes when the robot successfully reaches the target location or when the maximum time limit is exceeded. 
The success criterion is defined as the distance between the robot and the target location being less than a predefined threshold.

\textbf{Assets.} 
We include a rich collection of objects of varying types and sizes to ensure diversity in interaction properties. 
Specific categories include box-shaped objects, barrel-shaped objects, chairs, and doors, totaling approximately 50 distinct assets.
These assets are endowed with realistic physical attributes (e.g., mass, friction, collision meshes) to simulate authentic contact dynamics. 
Furthermore, all assets feature high-fidelity visual textures and materials to minimize the visual sim-to-real gap.

\textbf{Evaluation metrics.} We report three key metrics to assess navigation performance: success rate (SR), calculated as the percentage of trials where the robot successfully reaches the goal within the time limit; path length (PL), measuring the total distance traveled by the robot; and distance to goal (DTG), representing the final distance between the robot and the goal.

\textbf{Simulation Features.} Powered by Isaac Sim, the environment provides high-fidelity physics via GPU-accelerated PhysX and photorealistic rendering through RTX, ensuring realistic object interactions and minimizing visual discrepancies. 
Furthermore, RL-based policies trained in Isaac Lab~\cite{mittal2025isaac} can be directly deployed on our benchmark without modification, enabling seamless policy transfer.
These capabilities, along with support for diverse robot embodiments, effectively bridge the sim-to-real gap, ensuring that algorithms maintain robust performance in real-world deployments and demonstrating the practical value of our dataset.

\section{Simulations}
\label{sec:simulations}

In this section, we evaluate CoINS through comprehensive experiments.
The evaluation is divided into following parts:
(1) VLM reasoning evaluation, where we assess the reasoning and decision-making capabilities of InterNav-VLM on different embodiments;
(2) Interactive navigation in simulation, where we compare CoINS against baselines in the proposed interactive navigation benchmark;
(3) Ablation studies, where we analyze the contribution of each key component in CoINS;
(4) Generalization analysis, where we validate the generalization of CoINS to wheeled non-holonomic robots;
(5) Interactive behavior visualization, where we qualitatively demonstrate the interaction behaviors with diverse objects;
(6) Failure case analysis, where we summarize the primary causes of failure and discuss directions for future improvements.

\subsection{VLM Reasoning Evaluation}
\begin{figure*}[!t] 
    \centering
    \includegraphics[width=\linewidth]{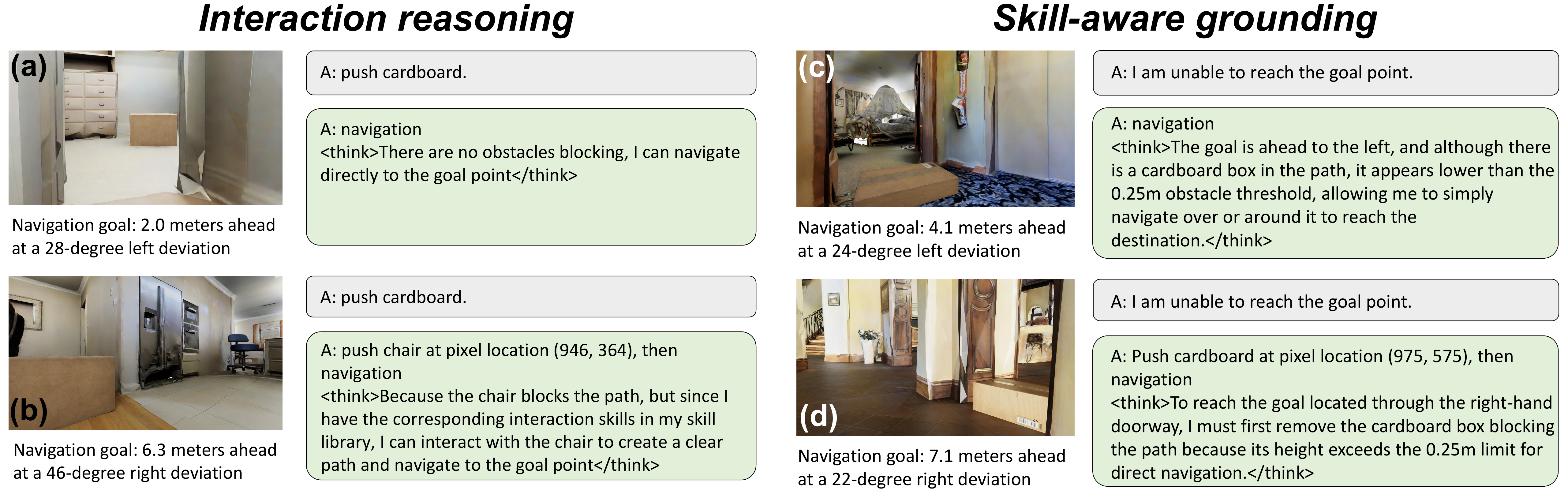} 
    \caption{Case study of InterNav-VLM reasoning. The response in the gray box is from the Qwen3-VL-8B-Instruct, while the response in the green box is from InterNav-VLM.
    Given the identical embodiment description and target goal, InterNav-VLM accurately identifies the necessary interaction and the target object, whereas the original VLM fails.}
    \label{Fig: VLM_case_study} 
    \end{figure*}

To validate the reasoning capabilities of InterNav-VLM, we evaluate its performance on the test set of VQA samples.
Specifically, we measure the model's accuracy in skill selection and object selection.
A prediction is considered correct only if the model selects the appropriate skill and, in cases requiring interaction, correctly identifies the target object.

We benchmark InterNav-VLM against its pre-trained counterpart, Qwen3-VL-8B-Instruct, and two representative proprietary models, GPT-4o~\cite{achiam2023gpt} and Gemini-2.5-Pro~\cite{team2023gemini}.
As illustrated in \Cref{tab:vlm_accuracy}, our model exhibits superior semantic understanding of the environment.
Quantitatively, our method achieves a significantly higher success rate on the test set, demonstrating that fine-tuning on domain-specific interactive navigation data is essential for robust decision-making in complex, cluttered environments.

\begin{table}[!h]
\centering
\small
\caption{Reasoning accuracy of different models on the test set.}
\label{tab:vlm_accuracy}
\begin{tabular}{lcc}
\toprule
\textbf{Model} & \textbf{Input} & \textbf{Accuracy (\%)} \\
\midrule
GPT-4o & RGB & 48.64 \\
Gemini-2.5-Pro & RGB & 58.34 \\
Qwen3-VL-8B-Instruct & RGB & 33.56 \\
\textbf{InterNav-VLM} & RGB & \textbf{78.35} \\
\bottomrule
\end{tabular}
\end{table}

To further analyze the skill-aware reasoning capability, we report the accuracy breakdown by two robot embodiments with different physical capabilities in \Cref{tab:vlm_accuracy_embodiment}, where wheeled robots can only navigate on planar ground without interaction skills, and legged manipulators can traverse over obstacles lower than 0.25m and interact with the environment.
The results show that InterNav-VLM achieves consistent improvements across both wheeled and legged manipulators, demonstrating its generalization ability to different skill capabilities.

\begin{table}[!h]
\centering
\small
\caption{Accuracy breakdown by robot type. The best result in each column is in bold.}
\label{tab:vlm_accuracy_embodiment}
\begin{tabular}{lcc}
\toprule
\textbf{Method} & \textbf{Wheeled} & \textbf{Legged Manipulator} \\
\midrule
GPT-4o & 58.89\% & 38.13\% \\
Gemini-2.5-Pro & 68.63\% & 45.09\% \\
Qwen3-VL-8B-Instruct & 45.44\% & 21.38\% \\
\textbf{InterNav-VLM} & \textbf{80.21\%} & \textbf{76.45\%} \\
\bottomrule
\end{tabular}
\end{table}

In addition to quantitative metrics, we present qualitative case studies in \Cref{Fig: VLM_case_study} to visually demonstrate the reasoning capabilities of InterNav-VLM.
The figure presents a comparison with the pre-trained VLM, Qwen3-VL-8B-Instruct, illustrating the improvements in high-level interaction reasoning and skill-awareness capability.
The left column in \Cref{Fig: VLM_case_study} demonstrates the ability of InterNav-VLM to reason about when and which object to interact with.
In \Cref{Fig: VLM_case_study}(a), where the target is directly reachable, the pre-trained VLM unnecessarily selects an interaction action, whereas InterNav-VLM correctly decides to navigate directly.
In \Cref{Fig: VLM_case_study}(b), the target is behind the chair on the right side; the pre-trained model erroneously chooses to interact with the cardboard on the left, which provides no benefit for reaching the goal. In contrast,  InterNav-VLM correctly identifies the right-side chair as the interaction target and provides precise coordinates, demonstrating its superior capability in selecting appropriate interaction objects.
The right column highlights the skill-awareness of InterNav-VLM, reflecting the model's understanding of embodied capabilities.
In \Cref{Fig: VLM_case_study}(c), InterNav-VLM correctly perceives that the obstacle's height falls within the robot's traversability threshold, thereby deciding to navigate directly.
Conversely, in \Cref{Fig: VLM_case_study}(d), confronting an obstacle that exceeds the robot's locomotive capabilities, the model appropriately determines that interaction is necessary to clear the path.

\begin{figure}[!t] 
\centering
\includegraphics[width=\linewidth]{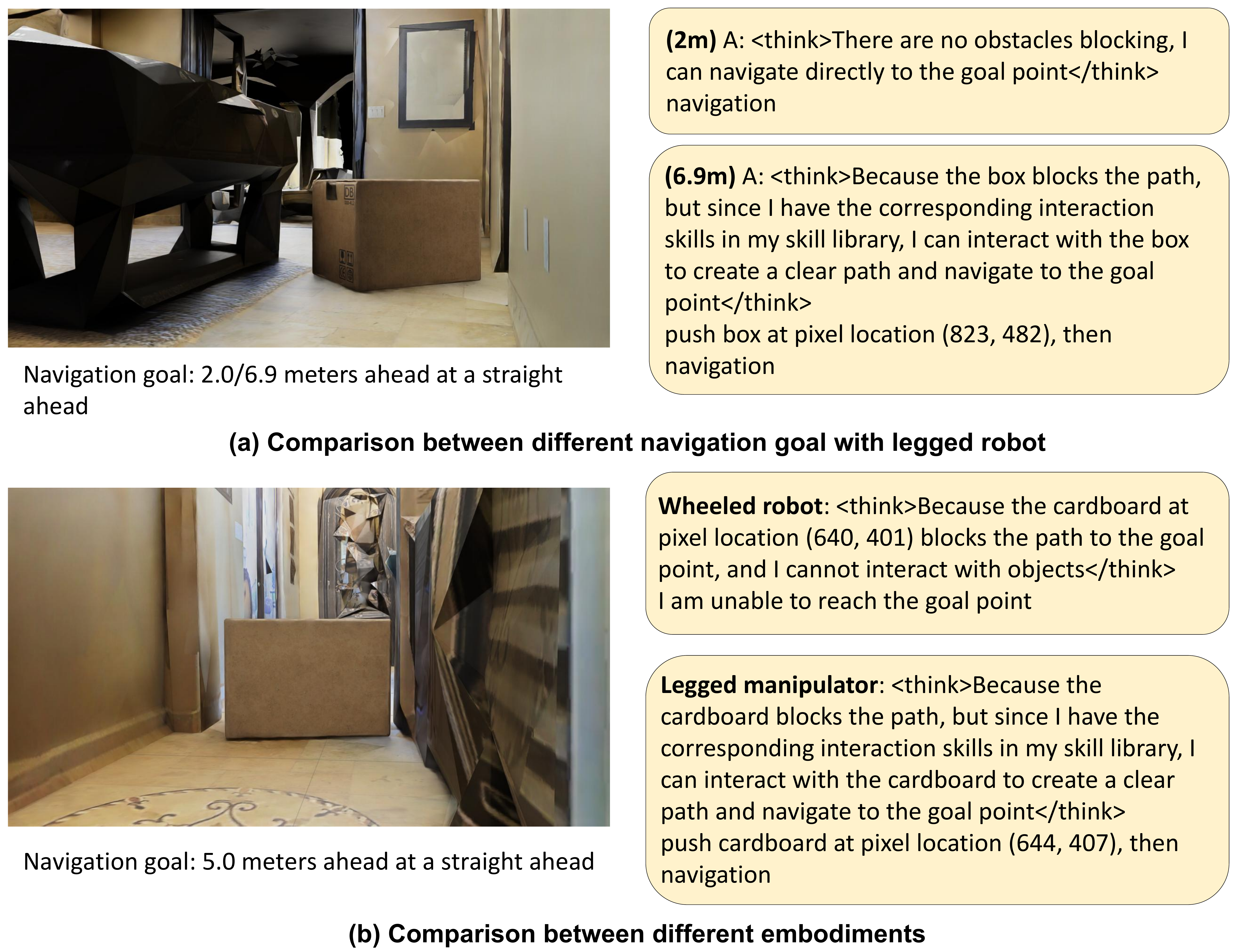} 
\caption{Case study of InterNav-VLM reasoning under different navigation goals and skill capabilities. }
\label{Fig: VLM_case_study2} 
\end{figure}

\Cref{Fig: VLM_case_study2} further demonstrates the reasoning capability of InterNav-VLM through two types of comparisons on different navigation goals and embodiments.
The first comparison involves the same legged robot embodiment tasked with different navigation goals: InterNav-VLM correctly selects direct navigation when the target is nearby and reachable, while accurately determining that interaction is required when the target is obstructed.
The second comparison presents the same navigation goal but with different embodiments. 
InterNav-VLM adapts its decision-making based on the available skill library of each embodiment, demonstrating its ability to reason about robot-specific capabilities and select appropriate actions accordingly.

\subsection{Interactive Navigation in Simulation}

We evaluate the interactive navigation performance on the proposed benchmark described in Section V.
The simulations are conducted using the Unitree Go2 quadruped robot equipped with an ARX5 manipulator arm as the robot platform.
We compare CoINS with the following baseline methods:
(1) \textbf{Art-planner}~\cite{wellhausen2023artplanner}: A sampling-based navigation method for quadruped robots that treats all objects as static obstacles and cannot interact with the environment.
(2) \textbf{IN-Sight}~\cite{schoch2024sight}: A traversability-based approach that combines global and local planners. It leverages global path planning for efficiency and employs local interaction strategies to handle obstacles.
(3) \textbf{IN-ArmPush}~\cite{bi2025interactive}: A search-based NAMO method that integrates reinforcement learning for arm-based object manipulation. Originally designed for known environments, we adapt it for unknown settings by replacing its ground-truth map dependency with local maps built from onboard sensors.
To isolate the impact of perception inaccuracies, we assume the robot has access to ground-truth object states for the RL policy observations.

\begin{table*}[t]
    \caption{Quantitative results of interactive navigation in simulation environments.}
    \label{tab:sim_results}
    \centering
    \resizebox{\textwidth}{!}{
    \begin{tabular}{l| *{3}{>{\centering\arraybackslash}p{1.4cm}} | *{3}{>{\centering\arraybackslash}p{1.4cm}} | *{3}{>{\centering\arraybackslash}p{1.4cm}} | *{3}{>{\centering\arraybackslash}p{1.4cm}} }
    \toprule
    \multirow{2}{*}{\textbf{Method}} & \multicolumn{3}{c|}{\textbf{Small Room}} & \multicolumn{3}{c|}{\textbf{Big Room}} & \multicolumn{3}{c|}{\textbf{Room-to-Room}} & \multicolumn{3}{c}{\textbf{Average}} \\
     & SR & PL (m)  & DTG (m) & SR & PL (m) & DTG (m) & SR & PL (m) & DTG (m) & SR & PL (m) & DTG (m) \\
    \midrule
    Art-planner & 0.24 & 12.43 & 3.16 & 0.18 & 16.43 & 4.46 & 0.12 & 24.07 & 8.08 & 0.18 & 17.64 & 5.23 \\
    IN-Sight & \textbf{0.88} & \textbf{5.88} & \textbf{0.29} & \underline{0.72} & \underline{10.21} & \underline{0.96} & \underline{0.32} & \textbf{14.69} & \underline{4.63} & \underline{0.64} & \textbf{10.26} & \underline{1.96} \\
    IN-ArmPush & 0.80 & 7.04 & 0.87 & 0.58 & 11.39 & 2.37 & 0.26 & 18.94 & 5.17 & 0.55 & 12.46 & 2.80 \\
    CoINS (Ours) & \underline{0.86} & \underline{6.07} & \underline{0.33} & \textbf{0.80} & \textbf{9.47} & \textbf{0.68} & \textbf{0.58} & \underline{15.43} & \textbf{2.57} & \textbf{0.75} & \underline{10.32} & \textbf{1.19} \\
    \bottomrule
    \end{tabular}}
    \end{table*}

We conducted a single test run for each of the 50 episodes in every scenario type and reported the average metrics. 
The quantitative results are summarized in \Cref{tab:sim_results}.
Art-planner, as a traditional navigation method, can only avoid obstacles without actively interacting with the environment. Consequently, it often encounters situations where it must take detours or fails to reach the goal when paths are blocked, resulting in poor performance across all metrics.
IN-Sight predicts traversability to identify interactable obstacles and employs a hierarchical planner to efficiently navigate through cluttered environments, achieving competitive performance in SR and PL. However, it does not utilize the manipulator arm and instead directly pushes through obstacles using the robot body, posing significant risks of damage to the robot's body and sensors in real-world deployment. Additionally, it struggles with irregularly-shaped obstacles and articulated objects like doors, often resulting in interaction failures.
IN-ArmPush learns manipulation strategies to move obstacles in confined spaces using the robotic arm, but it cannot handle diverse object categories and lacks the ability to adapt interaction modalities to different object types. 
Moreover, it adopts a two-stage approach that first samples target positions to relocate obstacles and then navigates, which demands high precision in manipulation and reduces overall navigation efficiency.

In contrast, our method achieves superior overall performance through two key contributions. 
First, InterNav-VLM intelligently determines when and what to interact with, leveraging its strong commonsense reasoning capability to select appropriate skills for different object types.
Second, our traversability-oriented manipulation policy efficiently handles scenarios where obstacles block the path, adapting to diverse object categories without requiring precise target placement.
Particularly in the room-to-room setting, our method demonstrates stronger capability in handling long-horizon tasks involving multiple interactions with diverse objects, including even opening doors, ensuring safe robot operation.

\begin{figure}[!t] 
\centering
\includegraphics[width=\linewidth]{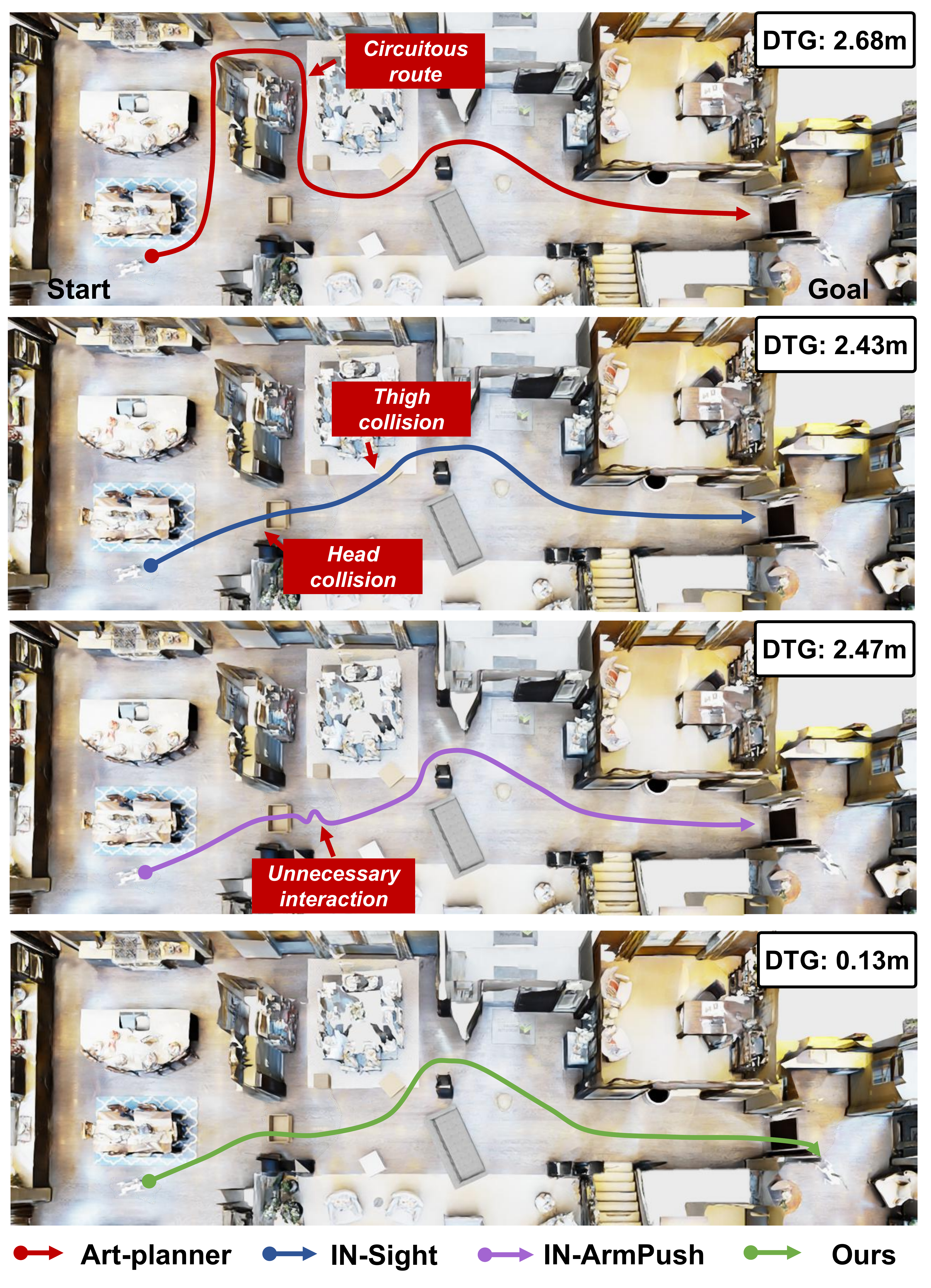} 
\caption{Qualitative results of interactive navigation in simulation environments. }
\label{Fig: qualitative_simulation} 
\end{figure}

\Cref{Fig: qualitative_simulation} visualizes the robot trajectories of all methods in a representative room-to-room scenario.
Art-planner, lacking interaction capabilities, is forced to take a detour when facing obstacles, significantly increasing the path length.
IN-Sight navigates efficiently in cluttered environments, but due to the lack of a whole-body control strategy, such as using the robotic arm to manipulate objects, it results in frequent collisions between the robot body and environmental obstacles, failing to meet safety requirements.
IN-ArmPush generates a reference path based on the current static map and interacts with objects obstructing this path, which often leads to unnecessary interactions.
For instance, if an obstacle is obstructing the path, IN-ArmPush will interact with it to clear the path, even if the obstacle can be easily circumvented.
Crucially, all three baseline methods fail to handle the closed door, resulting in large DTG values as they cannot reach the final target.
In contrast, our method leverages InterNav-VLM to intelligently determine the appropriate opportunity for interaction and utilizes diverse manipulation skills to handle various object types, including doors. 
This enables the robot to navigate efficiently and precisely to the goal while maintaining a low collision rate, outperforming all baselines.

\subsection{Ablation Studies}
To further analyze the effectiveness of our design choices, we conducted ablation studies on two key components of CoINS.
We sampled 20 episodes from each of the three scenario types, resulting in a total of 60 episodes for evaluation.
The ablation variants are:
\begin{itemize}
    \item \textbf{without fine-tuning:} Using the base Qwen3-VL model without domain-specific fine-tuning for interaction reasoning.
    \item \textbf{without traversability-oriented manipulation (TOM):} Replacing our traversability-oriented manipulation policy with a two-stage approach similar to IN-ArmPush, which first samples target positions to relocate obstacles and then navigates.
\end{itemize}

The results are summarized in \Cref{tab:ablation}.
Removing the fine-tuning leads to a significant drop in SR and a notable increase in DTG, demonstrating that domain-specific fine-tuning is critical for enabling InterNav-VLM to make accurate interaction decisions. The base VLM frequently fails to identify the correct interaction timing and target objects, leading to navigation failures.
Without the traversability-oriented manipulation policy, the two-stage approach results in lower SR and longer PL, confirming that requiring precise obstacle relocation before navigation reduces overall efficiency, whereas our TOM policy enables more flexible and efficient path clearance.
These results demonstrate that both components are essential for achieving robust interactive navigation performance.

\begin{table}[!h]
\centering
\footnotesize
\caption{Ablation study results.}
\label{tab:ablation}
\begin{tabular}{l *{3}{>{\centering\arraybackslash}p{1.4cm}} }
   \toprule
   \textbf{Method} & SR $\uparrow$ & PL (m) $\downarrow$ & DTG (m) $\downarrow$ \\
   \midrule
   w/o fine-tuning & 0.17 & 12.43 & 6.26 \\
   w/o TOM & 0.53 & 13.03 & 3.17 \\
   Ours & \textbf{0.68} & \textbf{11.56} & \textbf{1.42} \\
   \bottomrule
   \end{tabular}
\end{table}

\subsection{Adaptability to Different Embodiments}
We also conduct experiments to validate the adaptability of our method to different embodiments.
In addition to the Unitree Go2 quadruped robot equipped with an ARX5 manipulator arm used in the main experiments, we select TurtleBot3 as an alternative embodiment representing non-holonomic wheeled robots without manipulation capabilities.
We sample 30 episodes where traditional navigation methods can successfully reach the goal, ensuring that the scenarios are solvable without interaction for at least one embodiment.

We compare three configurations:
(1) \textbf{Art-planner + TurtleBot3}: Traditional navigation baseline on the wheeled platform.
(2) \textbf{CoINS + TurtleBot3}: Our method deployed on the wheeled platform, where InterNav-VLM reasons about navigation decisions based on the TurtleBot3's embodiment description without manipulation skills.
(3) \textbf{CoINS + Go2 w/ ARX5}: Our method deployed on the quadruped platform with manipulation capabilities.

The results are presented in \Cref{tab:cross_embodiment}.

\begin{table}[!h]
\centering
\footnotesize
\caption{Performance comparison on different embodiments.}
\label{tab:cross_embodiment}
\begin{tabular}{l *{3}{>{\centering\arraybackslash}p{1.4cm}} }
   \toprule
   \textbf{Method} & SR $\uparrow$ & PL (m) $\downarrow$ & DTG (m) $\downarrow$ \\
   \midrule
   Art-planner + TurtleBot3 & \textbf{0.93} & 8.96 & \textbf{0.21} \\
   CoINS + TurtleBot3 & \textbf{0.93} & 8.67 & 0.24 \\
   CoINS + Go2 w/ ARX5 & 0.87 & \textbf{6.74} & 0.43 \\
   \bottomrule
   \end{tabular}
\end{table}

Comparing the first two configurations (Art-planner + TurtleBot3 vs. CoINS + TurtleBot3), our method achieves comparable performance to the state-of-the-art navigation method on the same wheeled platform, with matching SR and similar PL and DTG. This demonstrates that our approach maintains strong obstacle avoidance and navigation capabilities even without interaction skills.
Comparing the latter two configurations (CoINS + TurtleBot3 vs. CoINS + Go2 w/ ARX5), our method achieves competitive performance across different embodiments. Notably, when the robot platform possesses manipulation capabilities, our method proactively interacts with the environment to find shorter paths, resulting in a significantly reduced PL. This behavior demonstrates that InterNav-VLM can intelligently adapt its strategy based on the available skills of each embodiment.

\subsection{Interaction Behavior Visualization}

\begin{figure}[!t] 
\centering
\includegraphics[width=\linewidth]{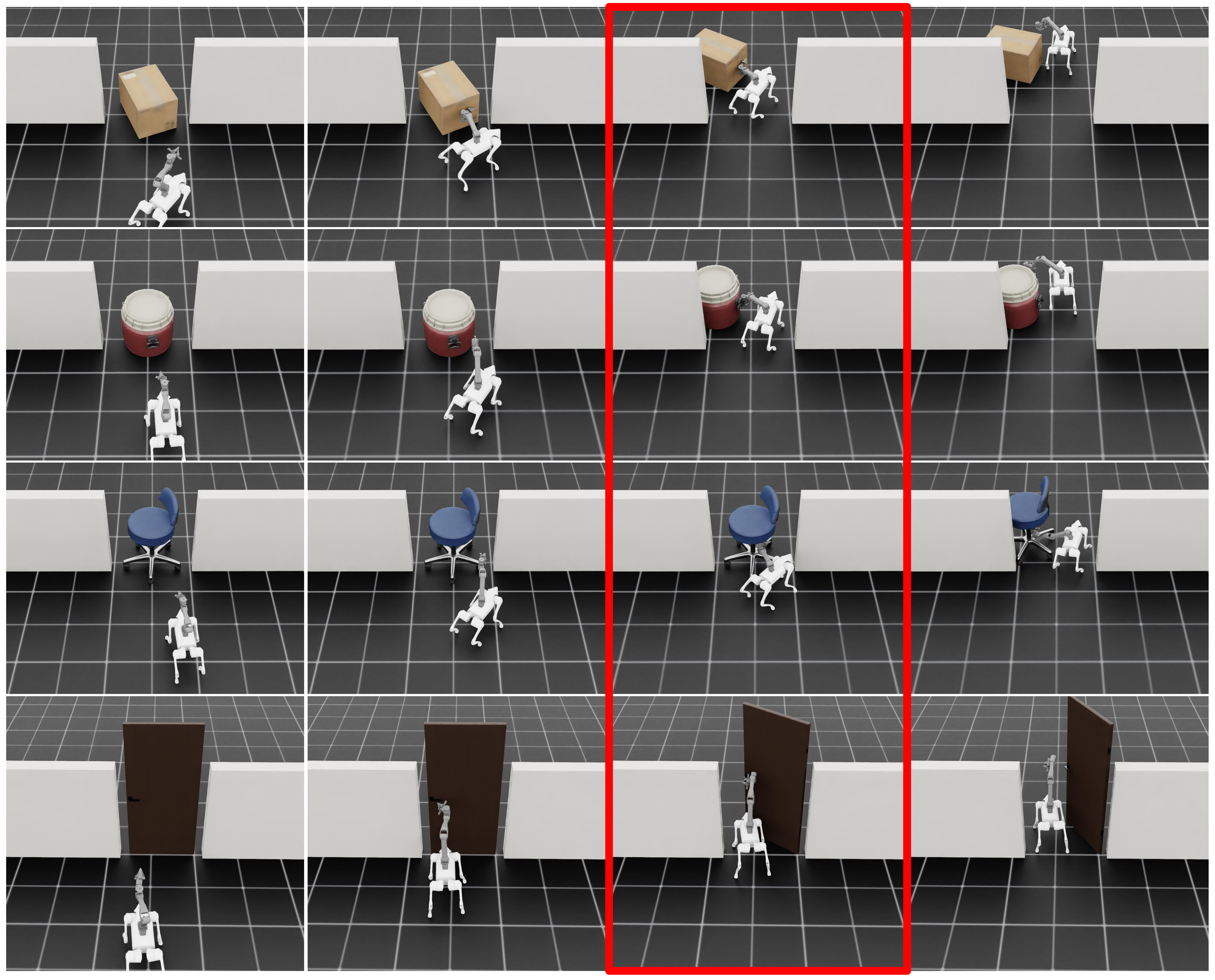} 
\caption{Visualization of interaction behaviors with different objects. 
The robot adapts its interaction strategy and contact points (highlighted in the red column) based on the object's shapes, such as pushing the edge of a cardboard or the gas lift cylinder of a chair, to efficiently clear a path.}
\label{Fig: interaction_behavior} 
\end{figure}

To provide a deeper insight into how our learned skills operate, we visualize the interaction behaviors in \Cref{Fig: interaction_behavior}.
The figure showcases the robot's interaction strategies with four distinct types of obstacles: a cardboard box, a barrel, an office chair, and a clockwise rotating door.
It can be observed that the robot employs lateral movements to interact with the objects, effectively clearing the path with minimal interaction displacement to reach the target goal.
This behavior perfectly aligns with the proposed traversability-oriented manipulation policy, demonstrating that the robot prioritizes path clearance over precise object placement.
Furthermore, the column highlighted in red illustrates that the interaction contact points vary significantly across different object categories to ensure effective manipulation.
For objects with simple geometries like the cardboard box and the barrel, the end-effector targets their edges to initiate the push.
In contrast, for the office chair, the robot specifically targets the gas lift cylinder to ensure stable interaction, leveraging the object's structure for effective displacement.
This adaptive contact selection demonstrates the versatility of our skill library in handling diverse object structures and physical properties.

\subsection{Failure Case Analysis}

\begin{figure}[!t] 
\centering
\includegraphics[width=\linewidth]{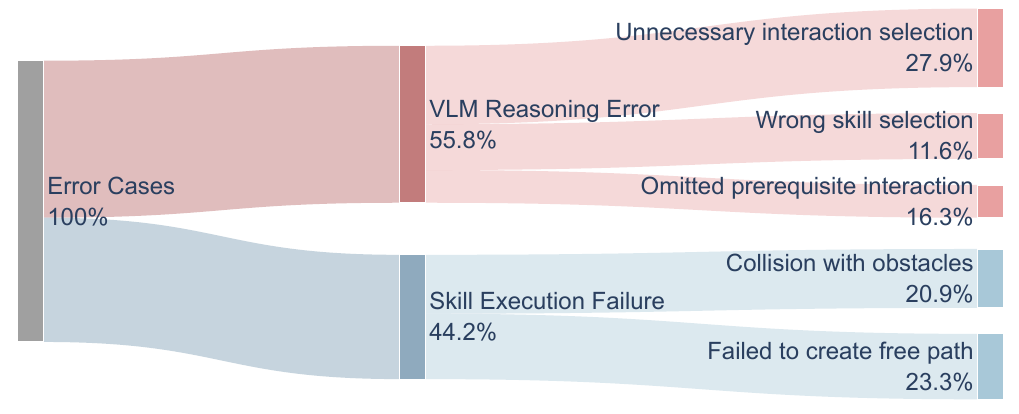} 
\caption{Distribution of failure cases.}
\label{Fig: failure_case} 
\end{figure}

To gain deeper insights into the limitations of our method, we systematically analyze all failure cases from the simulation experiments.
As illustrated in \Cref{Fig: failure_case}, the failure cases are categorized into two primary types: VLM reasoning errors and skill execution failures.
VLM reasoning errors consist of three subcategories:
(1) \textit{Unnecessary interaction selection}: The VLM incorrectly determines that interaction is required when the path is actually traversable, typically occurring in scenarios with narrow but passable gaps.
(2) \textit{Omitted prerequisite interaction}: The VLM fails to recognize that interaction is necessary when the path is blocked, leading to navigation failures as the planner cannot find a feasible route.
(3) \textit{Wrong skill selection}: The VLM selects an inappropriate skill for the object to interact with.
Skill execution failures are attributed to limitations in the low-level control policies.
(1) \textit{Collision with obstacles}: During interaction, the robot collides with surrounding obstacles in cluttered environments, causing instability or falls.
(2) \textit{Failed to create free path}: The manipulation skill contacts the target object but fails to displace it sufficiently to create a traversable path.

Future work will focus on addressing these limitations by enhancing the capability of InterNav-VLM to reason about spatial relationships and developing more robust manipulation policies.

\section{Real-World Experiments}

\subsection{Experimental Setup}

We conduct real-world experiments to validate the performance of our proposed framework in practical situations.
The robotic platform employed is the Unitree Go2 quadruped robot equipped with an ARX5 manipulator.
For environmental perception, the robot utilizes a Unitree L1 LiDAR to acquire terrain information.
Additionally, an onboard Intel RealSense D435i camera is mounted to capture ego-centric RGB images for VLM-based reasoning.

\begin{figure}[!t] 
\centering
\includegraphics[width=\linewidth]{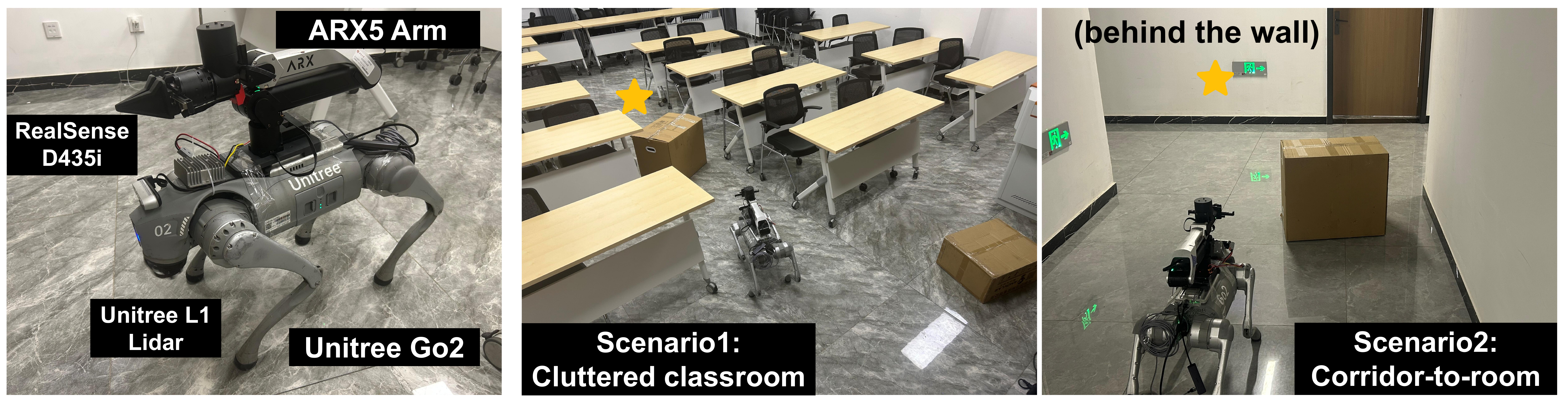} 
\caption{Real-world experimental setup, including the robot platform and the two test environments.}
\label{Fig: real_exp} 
\end{figure}

As illustrated in \Cref{Fig: real_exp}, we evaluate our method in two challenging scenarios:
(1) cluttered classroom: a confined environment filled with obstacles such as chairs and cardboard boxes. The robot's navigation goal is located on the other side of these obstacles, requiring it to navigate through the clutter.
(2) corridor-to-room: a scenario situated in a corridor where the target point is inside a classroom with a closed door. The robot must proactively interact with the door to open it and reach the destination.

\subsection{Qualitative Results}

\begin{figure*}[!t] 
    \centering
    \includegraphics[width=\linewidth]{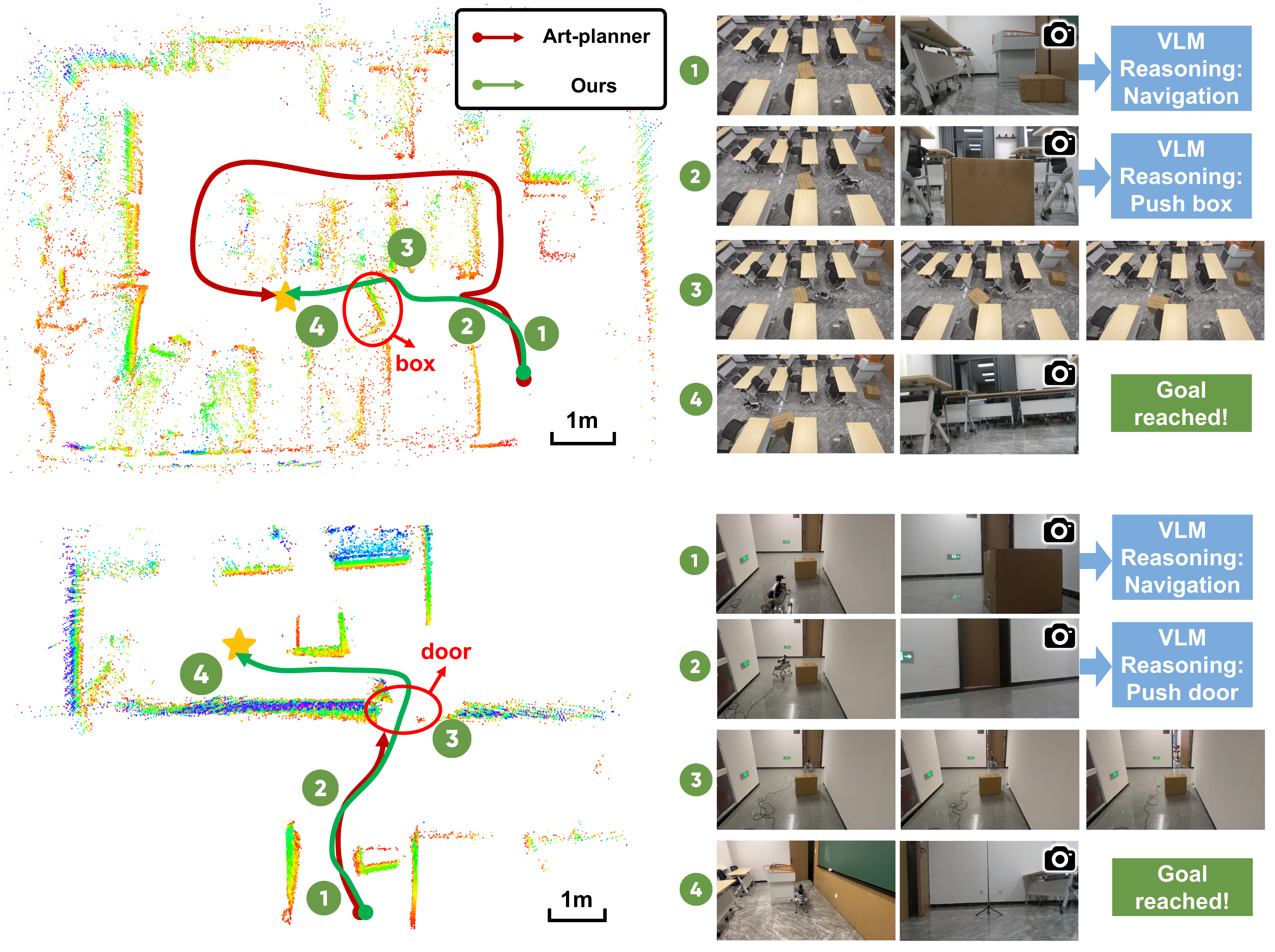} 
    \caption{Visualization of real-world experiments. The left side shows the point cloud and trajectory. The right side shows key frames corresponding to the numbers on our method's trajectory. Frames with a camera icon indicate the robot's ego-centric view.}
    \label{Fig: exp_vis} 
    \end{figure*}

\Cref{Fig: exp_vis} visualizes the specific navigation processes in both scenarios.
The left panel displays the real-time point cloud results and the robot's trajectory, where the green line represents the path generated by our method.
The red numbers on the trajectory correspond to the detailed moments shown on the right panel.
In the detailed views, images marked with a camera icon in the top-right corner represent the robot's ego-centric perspective, while others are third-person views provided to fully demonstrate the robot's behavior.

For the cluttered classroom scenario, the robot encounters a path blocked by the chaotic arrangement of desks and cardboard boxes.
In Step 1, the robot observes a cardboard box in its field of view. InterNav-VLM reasons that this object does not obstruct the path to the goal and decides to continue navigation.
Upon reaching Step 2, the robot encounters another box that blocks the main passageway. Recognizing this as an obstacle, the VLM selects the interaction skill to clear the path.
In Step 3, the robot executes the traversability-oriented manipulation policy, laterally pushing the box aside to efficiently create a gap.
Finally, in Step 4, the robot successfully navigates through the cleared path and reaches the target goal.

In the corridor-to-room scenario, a similar decision-making process is observed.
In Step 1, the VLM determines that the box in front does not hinder navigation and chooses to proceed.
However, in Step 2, the robot faces a closed door blocking access to the target room. The system correctly decides to interact with the door.
In Step 3, the robot executes the door opening skill, using its end-effector to push the door open.
Subsequently, the robot navigates through the doorway and reaches the destination inside the room.

In contrast, the baseline Art-planner, limited to passive obstacle avoidance, is forced to take significant detours in cluttered environments or fails completely when blocked by closed doors, highlighting the necessity of active interaction capabilities for navigation in realistic, unstructured settings.

The successful deployment in these two complex real-world scenarios demonstrates that InterNav-VLM can adaptively determine interaction timing and select appropriate skills based solely on RGB inputs.
It also validates the effective sim-to-real transfer of our diverse interaction skills.
By combining high-level semantic reasoning with robust low-level control, our method successfully accomplishes interactive navigation tasks in cluttered and cross-room environments, proving its applicability and robustness in real-world settings.

\subsection{Skill Execution Analysis}

\begin{figure}[!t] 
\centering
\includegraphics[width=\linewidth]{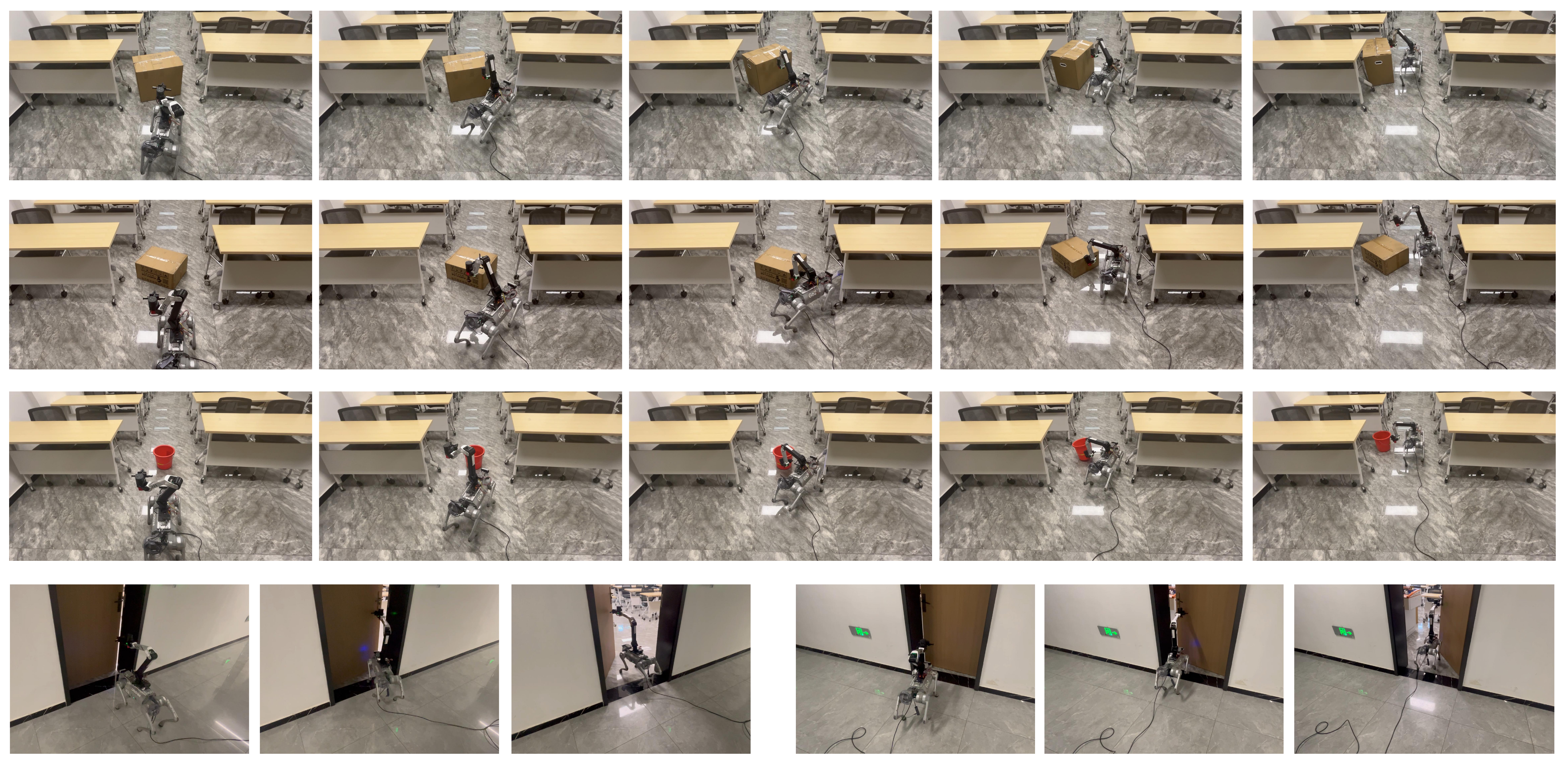} 
\caption{Real-world skill execution evaluation on diverse objects. The robot successfully adapts its manipulation strategy to different object shapes and door types.}
\label{Fig: exp_skill} 
\end{figure}

In addition to the navigation tasks, we explicitly evaluate the execution of individual interaction skills on real-world objects to verify their generalization capabilities, as shown in \Cref{Fig: exp_skill}.
We tested the robot's ability to interact with various obstacles, including two cardboard boxes of different dimensions ($60 \times 40 \times 50$ cm and $55 \times 45 \times 20$ cm), a small bucket, as well as left-hinged and right-hinged doors.

During object interaction, we observed that the manipulator's end-effector adaptively adjusts its position to target different contact points based on the object's size and shape.
This behavior confirms the robustness of our traversability-oriented manipulation policy in handling obstacles with varying geometries, validating its generalization capability across different object categories.
Furthermore, in door opening tasks, the robot successfully identified the hinge and handle positions for both left-hinged and right-hinged doors, executing the appropriate pushing strategy to open them effectively.
These results demonstrate that our learned skills can robustly transfer to real-world objects with diverse physical properties.

\section{Conclusion}

In this work, we presented CoINS, a hierarchical framework for embodiment-aware interactive navigation, addressing the limitations of traditional navigation methods in cluttered and unstructured environments.
We introduced InterNav-VLM, a fine-tuned vision-language model capable of reasoning about "when" interaction is necessary and "which" object to interact with, explicitly accounting for the robot's embodiment capabilities.
To support the execution of these decisions, we developed a traversability-oriented manipulation policy using reinforcement learning, which prioritizes path clearance over precise object rearrangement, enabling efficient interaction with diverse objects.
Furthermore, we established a comprehensive interactive navigation benchmark to systematically evaluate performance in scenarios where environmental modification is a prerequisite for success.
Extensive experiments in simulation and real-world environments demonstrated that our method significantly outperforms existing baselines, showcasing robust generalization across different robot embodiments and object categories.
Our approach successfully integrates high-level semantic reasoning with low-level robust control, enabling robots to autonomously modify their surroundings to navigate complex spaces.

Future work will focus on enhancing the model's 3D spatial understanding to enable more precise navigation and interaction in complex environments.
We also plan to explore the integration of memory modules into InterNav-VLM to support long-horizon reasoning in large-scale environments.
Additionally, we aim to investigate the deployment of our framework on diverse robot platforms, including humanoid robots, to further validate its cross-embodiment generalization.

{
\bibliographystyle{ieeetr}
\bibliography{ref}
}


\appendix

\subsection{Metric-Scale Scene Reconstruction}
\label{appendix: scene-reconstruction}

We reconstruct a metric-scale 3D point cloud from a single RGB image and subsequently aligns it to a canonical world coordinate system for downstream reasoning tasks. 
The pipeline consists of three stages: Metric-scale depth estimation, point cloud reconstruction, and geometric canonicalization.

\subsubsection{Metric-Scale Depth Estimation}
To recover dense 3D geometry from the input RGB image $\mathbf{I} \in \mathbb{R}^{H \times W \times 3}$, we employ a hybrid depth estimation strategy. 
While VGGT~\cite{wang2025vggt} excel at capturing high-fidelity structural details, they typically generate normalized relative depth maps lacking physical scale. 
Conversely, metric-aware models such as Map-Anything~\cite{keetha2025mapanything} can estimate metric information but often result in coarser reconstructions with lower accuracy. 
To simultaneously achieve high-frequency structural details and metric-scale accuracy, we integrate the strengths of both approaches to derive metric-scale accurate depth estimation. 

We first utilize VGGT to generate a high-quality relative depth map $\mathbf{D}_{rel}$ alongside the camera intrinsic and extrinsic parameters:
\begin{equation}
\mathbf{D}_{rel}, \mathbf{K}, \mathbf{T} = \text{VGGT}(\mathbf{I}),
\end{equation}
where $\mathbf{K} \in \mathbb{R}^{3 \times 3}$ is the intrinsic matrix and $\mathbf{T} \in SE(3)$ represents the camera pose in the world frame.
Subsequently, we leverage Map-Anything to predict a coarse metric depth map $\mathbf{D}_{met}$, which serves as a reference to resolve the scale ambiguity.
Specifically, we compute a global scale factor $s$ using robust statistics over the set of valid intersection pixels $\Omega$, defined as pixels where both depth maps have valid values:

\begin{equation}
s = \underset{(u,v) \in \Omega}{\text{median}} \left( \frac{\mathbf{D}_{met}(u,v)}{\mathbf{D}_{rel}(u,v)} \right).
\end{equation}
The final metric depth map $\mathbf{Z} \in \mathbb{R}^{H \times W}$ is obtained by rectifying the relative depth:

\begin{equation}
\mathbf{Z} = s \cdot \mathbf{D}_{rel}.
\end{equation}

\subsubsection{Point Cloud Reconstruction}
Given the metric depth $\mathbf{Z}$ and the estimated intrinsic matrix $\mathbf{K}$, we reconstruct the raw point cloud $\mathcal{P}_{cam}$ via inverse projection. The intrinsic matrix $\mathbf{K}$ is defined as:

\begin{equation}
\mathbf{K} = \begin{bmatrix} f_x & 0 & c_x \\ 0 & f_y & c_y \\ 0 & 0 & 1 \end{bmatrix},
\end{equation}
where $(f_x, f_y)$ represents the focal length and $(c_x, c_y)$ denotes the principal point. For every pixel $(u, v)$ with depth $z = \mathbf{Z}(u,v)$, the corresponding 3D coordinate $\mathbf{p}_{cam} = [x, y, z]^T$ in the camera coordinate frame is computed as:

\begin{equation}
x = \frac{(u - c_x) \cdot z}{f_x}, \quad y = \frac{(v - c_y) \cdot z}{f_y}.
\end{equation}
The resulting point cloud is defined as $\mathcal{P}_{cam} = \{ \mathbf{p}_{cam}^{(i)} \}_{i=1}^{N}$.

\subsubsection{Geometric Canonicalization with Plane Alignment}
To facilitate downstream path planning, we transform the point cloud $\mathcal{P}_{cam}$ from the camera coordinate system to a canonical world coordinate system, ensuring the ground plane is orthogonal to the vertical axis.
First, to ensure robust ground plane estimation in cluttered environments, we rely on semantic information. We employ Grounding DINO~\cite{liu2024grounding} to identify the ground region in the RGB image. The corresponding 3D points $\mathcal{P}_{ground} \subset \mathcal{P}_{cam}$ are obtained via back-projection.
Next, we apply the Random Sample Consensus (RANSAC) algorithm specifically to $\mathcal{P}_{ground}$ to fit a plane model parameterized by a normal vector $\mathbf{n} \in \mathbb{R}^3$ and a distance scalar $d$:

\begin{equation}
\mathbf{n}^T \mathbf{p} + d = 0.
\end{equation}
Based on the estimated ground normal $\mathbf{n}$, we construct a canonical world coordinate system where the ground is aligned with the $XZ$-plane ($Y=0$) and the $Y$-axis points vertically upwards.
We define the new basis vectors $\{ \mathbf{u}_x, \mathbf{u}_y, \mathbf{u}_z \}$, where the new vertical axis $\mathbf{u}_y$ is aligned with the ground normal $\mathbf{n}$ (ensuring it points upwards). The new lateral axis $\mathbf{u}_x$ is determined to be orthogonal to both $\mathbf{u}_y$ and the camera's forward vector $\mathbf{v}_{fwd} = [0, 0, 1]^T$. The forward axis $\mathbf{u}_z$ is derived via the cross product:
\begin{equation}
\mathbf{u}_y = \frac{\mathbf{n}}{\| \mathbf{n} \|}, \quad \mathbf{u}_x = \frac{\mathbf{u}_y \times \mathbf{v}_{fwd}}{\| \mathbf{u}_y \times \mathbf{v}_{fwd} \|}, \quad \mathbf{u}_z = \mathbf{u}_x \times \mathbf{u}_y.
\end{equation}
Then, the rotation matrix $\mathbf{R}_{align}$ is constructed to project the point cloud from the camera frame to the new canonical basis:
\begin{equation}
\mathbf{R}_{align} = \begin{bmatrix} \mathbf{u}_x^T \\ \mathbf{u}_y^T \\ \mathbf{u}_z^T \end{bmatrix}.
\end{equation}
To align the ground plane to zero height, we select a reference ground point $\mathbf{p}_{ground} \in \mathcal{P}_{ground}$ and compute the translation vector $\mathbf{t}_{align}$ to move it to the origin:
\begin{equation}
\mathbf{t}_{align} = -\mathbf{R}_{align} \mathbf{p}_{ground}.
\end{equation}
We then formulate the homogeneous transformation matrix $\mathbf{T}_{align} \in \mathbb{R}^{4 \times 4}$:
\begin{equation}
\mathbf{T}_{align} = \begin{bmatrix} \mathbf{R}_{align} & \mathbf{t}_{align} \\ \mathbf{0} & 1 \end{bmatrix}.
\end{equation}
Finally, the canonicalized point cloud $P$ is obtained by transforming each point $\mathbf{p}_{cam}$ using the homogeneous matrix:
\begin{equation}
\begin{bmatrix} {p} \\ 1 \end{bmatrix} = \mathbf{T}_{align} \begin{bmatrix} \mathbf{p}_{cam} \\ 1 \end{bmatrix}.
\end{equation}
Correspondingly, the camera extrinsic matrix $\mathbf{T}$ is updated by 
\begin{equation}
\mathbf{T}_{new} = \mathbf{T} \mathbf{T}_{align}^{-1},
\end{equation}

\subsection{Reward Design}
\label{appendix: reward-design}

We list the symbol definition and the reward terms of skills in the hierarchical control framework as follows, with $\phi(x):=\exp(-\frac{||x||^2}{0.25})$.

\begin{table}[H]\centering
\caption{Definition of symbols.}
\begin{tabular}{rl}
    \toprule
    \textbf{Symbol} & \textbf{Description}\\
     \midrule
     $\mathbf{q}_j$ & Joint positions \\
     $\dot{\mathbf{q}_j}$ & Joint velocities \\
     $\ddot{\mathbf{q}_j}$ & Joint accelerations \\
     ${\mathbf{q}^*_j}$ & Target joint positions \\
     ${\mathbf{q}^{*,pre}_j}$ & Last target joint positions\\
     $\boldsymbol{\tau}_j$ & Joint torques\\
     $\mathbf{v}_{b}$ & Base linear velocity\\
     $\boldsymbol{\omega}_{b}$ & Base angular velocity \\
     $\boldsymbol{x}_{b}$ & Base position\\
     $\boldsymbol{\theta}_{b}$ & Base heading\\
     $\boldsymbol{x}_{o}$ & Object pose\\
     $\mathbf{v}^*_{b}$ & Commanded base linear velocity \\
     $\boldsymbol{\omega}^*_{b}$ & Commanded base angular velocity \\
     $\boldsymbol{x}^*_{b}$ & Commanded base position\\
     $\boldsymbol{x}^*_{o}$ & Commanded object pose\\
     $\boldsymbol{x}_{ee}$ & End-effector position \\
     $\boldsymbol{x}^*_{ee}$ & Commanded end-effector position \\
     $\theta_{door}$ & Door opening angle \\
     $\theta^*_{door}$ & Target door opening angle \\
     $n_{c}$ & Number of collisions \\
     $\mathbf{t}_{air, f}$ & Feet air time \\
 \bottomrule
\end{tabular}
\label{table:nomenclature}
\end{table}

\subsubsection{Low-level whole-body controller}

The low-level controller tracks the commanded base velocity $\mathbf{v}^*_b$ and end-effector pose $\boldsymbol{x}^*_{ee}$ from high-level skills while maintaining robot stability.

\begin{table}[H]\centering
\footnotesize
\caption{Reward terms in low-level whole-body controller.}
\begin{tabular}{rcl}
    \toprule
     \textbf{Reward terms} & \textbf{Definition} & \textbf{Weight}\\
     \midrule
     Linear velocity tracking & $\phi(\mathbf{v}^*_{b,xy} - \mathbf{v}_{b,xy})$ & $1.0$ \\
     Angular velocity tracking & $\phi(\boldsymbol{\omega}^*_{b,z} - \boldsymbol{\omega}_{b,z})$ & $0.5$\\
     End-effector position tracking & $\phi(\boldsymbol{x}^*_{ee} - \boldsymbol{x}_{ee})$ & $1.0$ \\
     Linear velocity penalty & $-\mathbf{v}_{b,z}^2$ & $2.0$ \\
     Angular velocity penalty & $-||\boldsymbol{\omega}_{b,xy}||^2$ & $0.05$\\
     Joint torques & $-||\boldsymbol{\tau}_j||^2$ & $10^{-5}$\\
     Joint accelerations & $-||\ddot{\mathbf{q}_j}||^2$ & $2.5 \times 10^{-7}$ \\
     Action rate & $-||\mathbf{q}^*_j-\mathbf{q}^{*,pre}_j||^2$ & $0.01$ \\
     Collisions & $-n_{c}$ & $1.0$\\
     Feet air time & $ \sum_{f=0}^{4}(\mathbf{t}_{air, f} - 0.5)$ & $0.125$\\
     Foot Joint Position & Deviation from nominal (Legs) & $-0.7$ \\
     Feet Slide & Velocity of feet in contact & $-0.2$ \\
 \bottomrule\\
\end{tabular}
\label{table:rewards_lowlevel}
\end{table}

\subsubsection{Traversability-oriented manipulation skill}

This skill outputs base velocity commands and end-effector poses to navigate the robot to a target position $\boldsymbol{x}^*_b$ while using the manipulator to push aside obstructing objects.

\begin{table}[H]\centering
\footnotesize
\caption{Reward terms in traversability-oriented manipulation skill.}
\begin{tabular}{rcl}
    \toprule
      \textbf{Reward terms} & \textbf{Definition} & \textbf{Weight}\\
     \midrule
      \multicolumn{3}{l}{\textit{Navigation reward $r_{nav}$}} \\
      Position tracking & $\phi(\boldsymbol{x}^*_{b} - \boldsymbol{x}_{b})$ & $5.0$ \\
      Obstacle clearance & $\text{dist}(\boldsymbol{x}_{o}, \boldsymbol{x}_{b} \to \boldsymbol{x}^*_{b})$ & $1.0$ \\
      \midrule
      \multicolumn{3}{l}{\textit{Safety reward $r_{safe}$}} \\
      Body collision penalty & $-n_{c}$ & $10.0$\\
      Stability penalty & $-||\boldsymbol{\omega}_{b,xy}||^2$ & $0.1$\\
      \midrule
      \multicolumn{3}{l}{\textit{Efficiency reward $r_{eff}$}} \\
      Joint torques & $-||\boldsymbol{\tau}_j||^2$ & $10^{-5}$\\
      Action rate & $-||\mathbf{q}^*_j-\mathbf{q}^{*,pre}_j||^2$ & $0.01$ \\
 \bottomrule\\
\end{tabular}
\label{table:rewards_tom}
\end{table}

\subsubsection{Door opening skill}
This skill opens a door to a target angle $\theta^*_{door}$ while avoiding collisions between the robot body and the door structure.

\begin{table}[H]\centering
\footnotesize
\caption{Reward terms in door opening skill.}
\begin{tabular}{rcl}
    \toprule
      \textbf{Reward terms} & \textbf{Definition} & \textbf{Weight}\\
     \midrule
      Door angle tracking & $\phi(\theta^*_{door} - \theta_{door})$ & $5.0$ \\
      End-effector to handle & $-||\boldsymbol{x}_{ee} - \boldsymbol{x}_{handle}||$ & $1.0$ \\
      Body collision penalty & $-n_{c}$ & $10.0$\\
      Stability penalty & $-||\boldsymbol{\omega}_{b,xy}||^2$ & $0.1$\\
      Joint torques & $-||\boldsymbol{\tau}_j||^2$ & $10^{-5}$\\
 \bottomrule\\
\end{tabular}
\label{table:rewards_door}
\end{table}

\end{document}